\documentclass{article}

\usepackage{arxiv}

\usepackage[utf8]{inputenc} 
\usepackage[T1]{fontenc}    
\usepackage{hyperref}       
\usepackage{url}            
\usepackage{booktabs}       
\usepackage{amsfonts,amsmath,amssymb}       
\usepackage{nicefrac}       
\usepackage{microtype}      
\usepackage{graphicx}
\usepackage{caption}

\usepackage{color, colortbl}

\usepackage{amsthm}
\usepackage{xcolor}
\usepackage{multirow}
\usepackage{subcaption}
\usepackage{tabularx,adjustbox}

\newcommand{\ie}{\textit{i.e.,}}
\newcommand{\eg}{\textit{e.g.,}}
\newcommand{\commenttxt}[1]{}

\newcommand{\mybar}{\kern1pt\rule[-\dp\strutbox]{.8pt}{\baselineskip}\kern1pt}
\newcommand{\expt}[1]{\ensuremath{\pmb{\mathcal{E}}\pmb{#1}}}

\renewcommand{\mytagline}{} 
\renewcommand{\mydate}{June 26, 2026} 

\theoremstyle{definition}
\newtheorem{definition}{Definition}[section]

\title{
    Consensus Clustering of Free-Viewing Gaze Data: New Insights into Human-Information Interaction\thanks{Codebase: \url{https://github.com/GVCL/EnsembleGaze}}}
\author{
  Beryl Gnanaraj\textsuperscript{1},  
  Jaya Sreevalsan-Nair\textsuperscript{1},\thanks{\texttt{jnair@iiitb.ac.in}}
  Saqib Alam Ansari\textsuperscript{1},
  Maanasa Rajaraman\textsuperscript{2} 
  \\
  \textsuperscript{1}Graphics-Visualization-Computing Lab,\\
  International Institute of Information Technology Bangalore, Karnataka 560100, India. \\
  \texttt{http://www.iiitb.ac.in/gvcl} \\
  \textsuperscript{2}PSG College of Technology, Coimbatore, Tamin Nadu, India. \\
}

\begin{document}
\maketitle

\begin{abstract}
Free-viewing gaze data provides a rich, task-free window into human visual attention. Conventional exploratory data analysis (EDA) of the data provides user attention patterns through fixations, and areas of interest (AOIs). However, despite the richness of this gaze data, its human-information interaction (HII) patterns are understudied. We address this gap using consensus clustering of gaze data with respect to users and stimulus characteristics. We present a novel end-to-end unsupervised ensemble learning system for consensus clustering of free-viewing gaze datasets, EnsembleGaze. With a goal of characterizing the user behavior and image stimulus type, we propose a feature engineering step based on statistical descriptors of fixation-based feature distributions. EnsembleGaze involves consensus voting of selected clustering methods implemented on the feature vector to compute the co-association matrix. Using the separate consensus clustering of users and stimuli as baseline, we further propose two high-dimensional clustering strategies for determining gaze clusters based on joint user and image characterization. They are consensus subspace clustering and spectral biclustering. Clustering performance is evaluated using selected standard metrics, and are further interpreted through image-level properties, such as dominant color, luminance intensity, and object properties. Our system provides a replicable, dataset-agnostic methodology for the unsupervised analysis of fixation behavior in scene perception research. Our results show that image stimuli groupings are highly consistent across methods, reflecting a robust ambient-versus-focal viewing mode distinction, whereas user groupings are image-context-dependent, a structure that only biclustering and the two-step conditional approaches are architecturally capable of recovering. Testing on the publicly available datasets, MIT1003 and EMOd, revealed dataset-specific patterns, with each offering complementary insights through distinct clustering strategies.
\end{abstract}


\section{Introduction}\label{sec:introduction}
Gaze data, captured through eye-tracking technology, provides rich information about user behavior, revealing where and for how long individuals focus their attention~\cite{yarbus1967eye, just1980theory, rayner1998eye}. This data is invaluable across various domains, such as psychology, marketing, and human-computer interaction, for understanding cognitive processes and preferences. However, traditional exploratory data analysis (EDA) of gaze data is often limited to specific data characteristics, such as fixation duration or saccade patterns~\cite{velichkovsky2002towards, unema2005time, naqshbandi2016automatic, yoo2018gaze}, without exploring aggregated analysis of user behavior and human-information interaction (HII). Here, HII refers to the interactions between the human and the images provided in the free-viewing task. User's attentional behavior analysis is different from the conventional saliency prediction using image characteristics~\cite{itti1998model, bruce2005saliency, judd2009learning, vig2014large, bruno2020multi} or visual fixation prediction in images~\cite{yue2021human, wu2025evaluating}, \ie~human saliency maps or fixation density maps, respectively. In the case of free-viewing gaze data from image stimuli, there is a gap in studies integrating user interaction and contextual visual features. Especially in vision tasks, these features have the potential for insights into the image characteristics that influence user behavior.

The state-of-the-art clustering of gaze data using machine learning (ML) provides EDA and reveals common gaze patterns across participants, areas of interest (AOIs), and gaze targets within visual stimuli~\cite{eraslan2016scanpath, kumar2018visual, kumar2019clustered, selim2024review}. In addition to reducing volume of raw gaze data to representative groups, gaze clusters reveals latent patterns in gaze and behavior. However, the clustering criteria in gaze data have not expanded in recent times despite the richness of the data. For instance, clusters based on image stimuli properties or users with similar attentional profiles can predict the gaze data for an unseen image or user, based on cluster membership. At the same time, clusters based on understudied criteria using unsupervised methods requires careful evaluation of effectiveness of existing clustering methods. Relying on specific methods leads to biases, which are difficult to detect in the absence of ground truth. Hence, to reduce modeling bias and to overcome uncertainty in validation, we propose the use of unsupervised ensemble learning, for which, we propose the approach of consensus voting for implementation. 

\begin{definition}
  \textbf{Unsupervised Ensemble Learning for Clustering} is a machine learning paradigm that integrates multiple learning models for clustering, and makes a final decision of data partition using a meta-learning model.
\end{definition}

\begin{definition}
  \textbf{Consensus Voting} is a method of using majority or average value of the predictions from an ensemble of machine learning models, which is referred to as hard voting or soft voting, respectively. This method should be seen as a reduction process of determining a final outcome from a pool of outcomes.
\end{definition}

Matrix analysis has been used for gaze data visualization for comparing multiple user metrics. One of the commonly used matrix is the transition matrix, where the probability of each AOI transition is computed from the frequency of the user's gaze shifting from the source AOI to the destination~\cite{ellis1986statistical, ponsoda1995probability, blascheck2014state, krejtz2015gaze, blascheck2017visualization}. Another matrix that is used in gaze data clustering is the similarity matrix~\cite{brandt1997spontaneous, jarodzka2010vector, kumar2016multi, kumar2019clustered}. These similarity matrices are computed using distance matrices built on pair-wise scanpaths. In this work, we explore the use of a less explored matrix, namely the co-association matrix~\cite{fred2002data, fred2005combining}, for unsupervised ensemble clustering of gaze data. These matrices help in understanding cluster dynamics among users and stimuli, and is used in a meta-learning model to obtain the final clustering. 

\begin{definition}
  \textbf{Co-association Matrix} is a data representation, specifically for a grouping problem, such as clustering or classification, where a symmetric matrix represents data items in both rows and columns in the same permutation, with its elements indicating the frequency or probability at which pairwise data items belong to the same group.

  Co-association matrix of a base clustering in an ensemble is a binary matrix, and the consensus voting of the base clusterings gives either a probability or frequency, upon averaging or summing, respectively.
\end{definition}

The richness of free-viewing gaze data indicates we can go beyond simple user- or stimuli-clusters, and perform high-dimensional clustering of (user, stimulus) combinations~\cite{cheng2000biclustering, getz2000coupled, dhillon2003information, kluger2003spectral}. Thus, we propose high-dimensional clustering of gaze data to understand the degree of influence of user behavior and image stimulus on the gaze data. As downstream steps, high-dimensional clustering can also help to more accurately predict or impute gaze data, using cluster representative values. 

\begin{definition}
  \textbf{High-dimensional Clustering} is a clustering technique of data represented in a high-dimensional space, where feature vector representation is inadequate for meaningful clustering. For instance, a data matrix of feature vectors is represented as a 3D data tensor consisting of two distinct entity types in two dimensions, and the feature vector in the third dimension, where each cluster involves tuples of entity values from the first two dimensions.

  If the data tensor is projected to a data matrix of the two dimensions, then high-dimensional clustering identifies appropriate clusters in submatrices, or similar ways of simultaneously including the rows and columns. This is different from one-way clustering, which involves clustering entire rows exclusively, or in certain cases, entire columns exclusively.
\end{definition}

\begin{figure*}[t]
  \centering
  \includegraphics[width=1\linewidth]{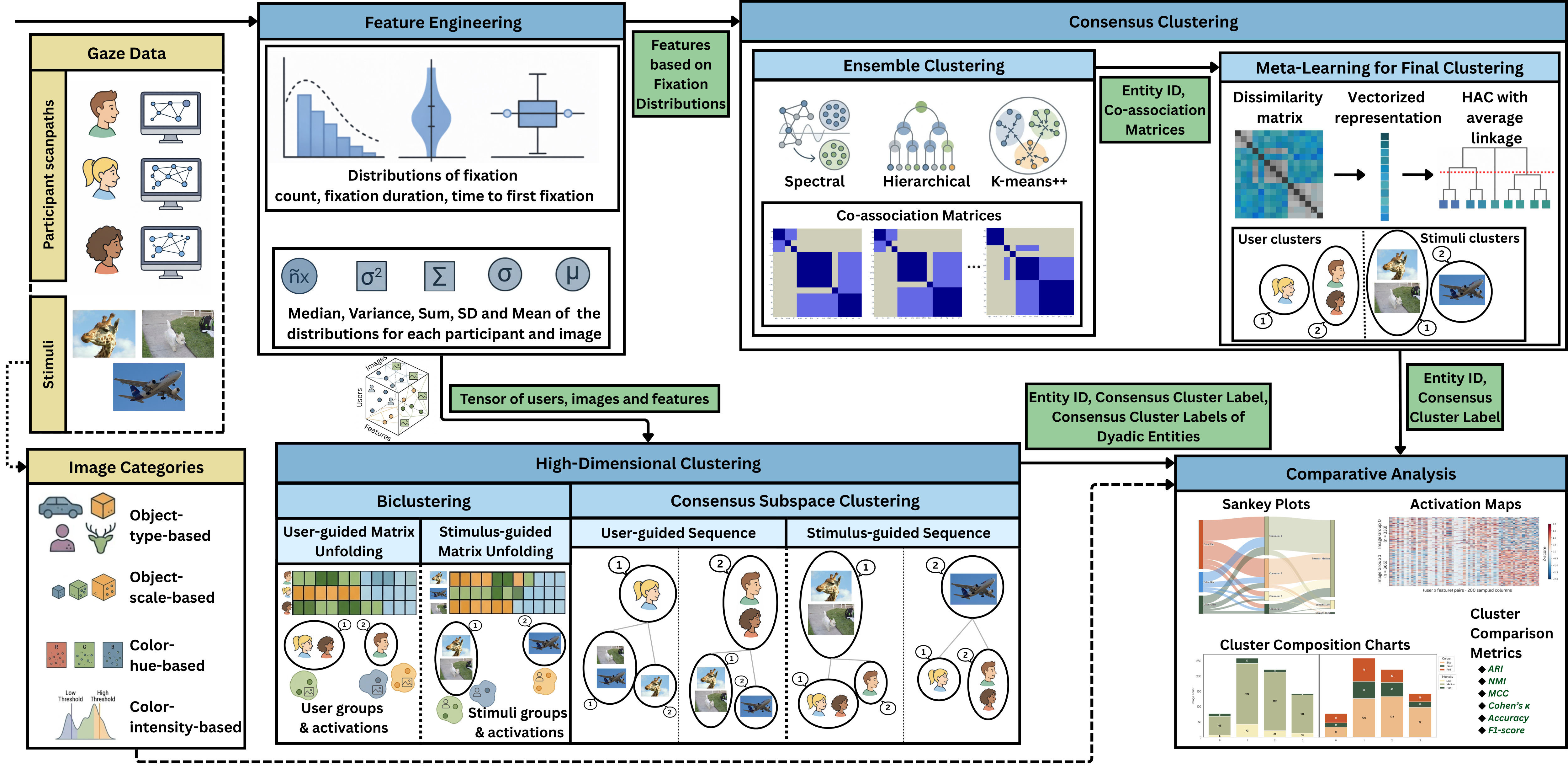}
  \caption{EnsembleGaze provides an end-to-end unsupervised ensemble clustering of free-viewing gaze data, involving feature engineering and consensus clustering. Consensus clustering involves co-association matrix computation and processing. Further, high-dimensional clustering is done using the consensus subspace clustering and biclustering modules. High-dimensional clustering outcomes are compared against the separate consensus clustering ones to determine the optimal clustering.}
  \label{fig:graphical_abstract}
\end{figure*}

To determine the optimal clustering outcome from the consensus clusters as well as high-dimensional clusters of gaze data, we design a novel end-to-end system called EnsembleGaze, which incorporates feature engineering, consensus clustering, and high-dimensional clustering, as shown in Figure~\ref{fig:graphical_abstract}. EnsembleGaze has been tested using two benchmark free-viewing gaze datasets, namely, MIT1003~\cite{judd2009learning} and EMOd~\cite{fan2018emotional}.

The contributions of our work are:
\begin{itemize}
\item Consensus clustering of gaze data separately based on image properties and user-behavior patterns for unsupervised ensemble learning.
\item High-dimensional clustering strategies adapted for integrating image and user-behavior information to identify their influence on the gaze data.
\item An end-to-end unsupervised ensemble learning system, EnsembleGaze, for clustering free-viewing gaze data.
\end{itemize}

\section{Background}\label{sec:background}
We discuss the state of the art on gaze clustering, feature engineering, and validation of clustering tasks.

\subsection{Clustering of Gaze Data}
Unsupervised clustering of gaze data has been widely studied in past literature~\cite{latimer1988eye, santella2004robust}. One prominent line of research focuses on identifying the most optimal and generalized clustering methods across various gaze domains~\cite{goldberg2010scanpath, eraslan2016scanpath, kumar2019clustered}. Concurrently, another critical branch of work addresses validation of clustering outcomes using statistical measures, in the absence of ground truth~\cite{kumar2018visual}. 

In reading research, the $k$-means algorithm has been used to infer reading patterns within a sample of participants reading scientific literature~\cite{lin2021modeling}, and group the users to fit into the three known reading patterns namely, reading through, re-reading and re-inspecting. It is not fully automated, as the gaze data was manually classified by visually analyzing reading patterns of each participant on each page. Similarly, depression recognition using gaze data~\cite{yang2022clustering} involves use of fixation-clusters to identify the Region-of-Interest (ROI) on image stimuli, which aids in classification of participants as depressed or healthy, while ASD identification~\cite{shi2024exploring} is done using clustering of gaze points to identify data-driven AOIs. Spatial and temporal features of gaze points for intent recognition are also gathered using ROIs~\cite{yoo2018gaze}. Similarly, the $k$-means and density-based clustering algorithms have been used for ROI identification in question answering tasks~\cite{naqshbandi2016automatic}. These works aim to identify fixations using gaze-point clusters~\cite{siradj2025towards}. However, in-depth analysis of user-, stimuli-, or (user, stimuli)-clusters is not studied in any of these works, and image properties are not correlated to the gaze behavior. 

A multi-metric clustering of gaze metrics used to identify similar participant group~\cite{kumar2016multi} has shown that using a single metric for identifying user group is challenging, due to the effect that the dataset has on the metric, leading to a multi-metric approach with weighted scoring of each metric for overall grouping. This work has been expanded to use dissimilarity matrix between pairwise users for each of the 16 metrics used~\cite{kumar2018visual}. The metrics included average measures of fixation and saccade features, numerical values such as $\kappa$ coefficient~\cite{krejtz2016discerning}, number of fixations and saccades, rates of fixations and saccades, and finally, distribution features based features on the gaze position $(x, y)$ on the screen, such as skew, standard deviation and kurtosis. The novelty of the work is in enhancing and cross-validating feature selection for clustering using visualizations, such as parallel coordinate and matrix plots. However, the visualizations were not scalable owing to the \textit{visual clutter} from fitting sixteen metrics for each user pair within a dissimilarity matrix. Further, this work expanded on effective matrix visualizations using various scanpath distance matrices to group the participants, and matrix reordering to find the user groups from the visualizations directly~\cite{kumar2019clustered}. Visualizations have also been used to show the saliency features of each image, prior to conducting eye-tracking experiments \cite{yoo2021saliency}. Co-association matrix visualization in a dashboard~\cite{gnanaraj2023eyeexplore} presents a novel method to look at similarity in user groups and images, by using consensus clustering, based on gaze features across each user and image. However, these works lacked validation using empirical measures. 

For high-dimensional clustering, biclustering~\cite{kluger2003spectral} was extensively employed in gene--condition analysis to identify subsets of genes that exhibit coordinated behavior under specific experimental conditions. Beyond this domain, biclustering serves as a general-purpose analytical approach for uncovering localized structures within multidimensional datasets across multiple interacting dimensions. Building upon approaches proposed in gene-expression analysis, coupled two-way clustering~\cite{getz2000coupled} identifies subsets of genes and samples such that clustering one dimension using the other yields stable and statistically significant partitions. Extending beyond bioinformatics, this methodology can also be adapted to eye-tracking research for identifying meaningful relationships between users, stimuli, and gaze behavior patterns.

Closely related to biclustering, subspace clustering~\cite{battaglia2024co} arises in the broader context of clustering high-dimensional data, in which the goal is to find clusters that exist in different subsets of the feature dimensions rather than in the full feature space. While biclustering simultaneously partitions both rows and columns to reveal localized co-expression patterns, subspace clustering relaxes the requirement for a global feature space and instead allows each cluster to inhabit its own low-dimensional subspace. The output of a subspace clustering algorithm is therefore a partition of the data points (rows) together with a basis for the subspace to which each cluster belongs, making the subspace itself an explicit part of the solution rather than merely a byproduct.

Despite the growing application of clustering techniques in gaze and eye-tracking analysis, the ideas of unsupervised ensemble clustering, consensus voting, biclustering, and consensus subspace clustering for gaze data are unexplored. To the best of our knowledge, these approaches have not previously been investigated in the context of gaze clustering. Motivated by the success of such methods in other multidimensional analytical domains, this study introduces and evaluates the applicability of these methods on gaze data, thereby establishing a novel direction for eye-tracking research.

\subsection{Feature Engineering for Gaze Data}\label{sec:lit_feature_engineering}
Different feature types have been used in clustering based on downstream applications, for instance, 2D static stimuli-based features are effective for ROI identification~\cite{yoo2018gaze, naqshbandi2016automatic, lin2021modeling, yang2022clustering, siradj2025towards}.  A set of sixteen features, including the eye-movement-based, numerical and spatial ones, has been effective in user clustering~\cite{kumar2016multi, kumar2018visual, kumar2019clustered}. These features can be used entirely or selectively, after visual evaluation of their correlations and negative associations. The derived numerosity measures, such as $\kappa$ coefficient, fixation rate, saccade rate, scanpath length, have been experimentally shown to be collinear with other fixation-based metrics that are included in the computations. In addition, fixations and saccades are simultaneously used in the feature set, which may have negative correlations with each other. 

Fixation features are known to emulate attention metrics. While saccades capture a shift in attention from one concept to another, fixations consider the area at which the gaze is localized for a period of time to be the only one being cognitively processed~\cite{just1980theory, rayner1998eye}. There are broadly two types of fixation features used in analyses~\cite{mahanama2022eye}, namely, fixation count and fixation duration. In reading research, additional fixation measures, like fixation line number, fixation duration, ROI markings of the fixations on each page are used~\cite{just1980theory, rayner1998eye, lin2021modeling}. Other fixation-based features like time-to-first fixation, fixation rate, on-target fixation, fixation spatial density have been explored in other studies~\cite{jacob2003eye, pieters2004attention}. In each fixation feature, a frequency distribution of the fixation is considered and additional features, like the mean, median, total, variance, standard deviation, 5$^{th}$ and 95$^{th}$ percentile have been considered in other studies~\cite{rigas2018study, gobel2018unsupervised, tarnowski2020eye, vortmann2021imaging}. 

The features are also classified as spatial and temporal, based on their characteristics. The spatial features included the $(x, y)$ gaze point, along with fixation length and duration, and the temporal ones included timestamps. Using simply the spatial features~\cite{naqshbandi2016automatic, yoo2018gaze, yang2022clustering}, clustering of gaze data aids in identifying ROIs. The temporal data helps in finding consistent user behavior patterns~\cite{rayner1998eye, lin2021modeling}. A classical user study has shown that higher fixation duration indicates higher cognitive processing, especially for reading tasks~\cite{just1976eye, just1980theory, rayner1998eye, rayner2009eye}. The "eye-mind hypothesis"~\cite{just1980theory} states that, the user's cognitive resources are dedicated to the part of the screen (or stimulus) that is currently being gazed at. This hypothesis is widely used for cognitive load estimation studies. There are various eye gaze features that have been explored with regards to understanding user's behavior and attention. The ones that can be directly computed from the gaze estimation module in eye trackers are fixation-based and saccade-based features. While fixation-based features serve the purpose of characterizing user attention to one area of a stimulus, saccade features, on the other hand, characterize a shift in attention from one ROI to another, and hence could be considered in the context of attention shifts. In image exploration studies~\cite{itti1998model, krejtz2016discerning}, higher saccade amplitudes and longer fixation duration correspond to ambient attention for visual scene exploration and focal attention for ROIs, respectively.

Since our work focuses on patterns in cognitive processing of areas of attention, we consider the fixation-based features, as per the eye-mind hypothesis. 

\subsection{Cluster Validation for Gaze Data}
Since clustering is a type of statistical learning method, the statistical way to estimate goodness-of-fit is by using the Fisher statistic $F$, \ie~between-group variance divided by the within-group variance. A large $F$ implies distinct or clear clusters, \ie~presence of highly discriminative features for clustering, which is not random. Similar metrics have been used in research, such as Cohen's $\kappa$, Matthew's Correlation Coefficient ($MCC$), Accuracy ($A$), and F1-score ($F_1$). 

A large number of gaze data analysis studies has not validated the goodness-of-fit of clustering methods using standard metrics~\cite{kumar2016multi, kumar2018visual, kumar2019clustered, yoo2018gaze, naqshbandi2016automatic}. The t-Stochastic Neighbor Embedding (t-SNE) is a popular visual validation method used on clustering outcomes~\cite{gobel2018unsupervised}. Similar visualizations have been used for validating gaze clusters~\cite{kumar2016multi, kumar2019clustered, gobel2018unsupervised}.  In a related work, as a rare example, clustering performance metrics have been used for evaluating classifiers used for ROI identification along with clustering~\cite{yang2022clustering}. These metrics included the widely used ones, such as precision, recall, accuracy, F1-score and Area under Curve (AUC). Estimating the best clustering method for gaze data is considered to be an open problem~\cite{kumar2018visual}, with a need for appropriate evaluation methods to measure the goodness-of-fit. While visualizations aid in providing a first-cut analysis of the results, thorough evaluation using statistical measures is necessary.

Thus, the state of the art shows the gaps in standardizing validation as a part of gaze data analysis, when clustering algorithms were used, the goodness-of-fit of the clustering algorithms were not measured empirically. Hence, in our study, we calculate metrics to measure the fit of clustering results, in addition to using the metrics for comparison across clustering methods. 

\section{Methodology}\label{sec:methodology}
In order to empirically obtain optimal features for unsupervised clustering, we run corresponding experiments with varied parameters in the feature engineering and clustering steps. The outcomes of the experiments are either user- or image-clusters. The workflow of our unsupervised ensemble clustering system, EnsembleGaze, in Figure~\ref{fig:graphical_abstract} involves ``Gaze Data'' as the input, and six processing steps, namely, ``Feature Engineering,'' ``Clustering,'' ``Consensus Clustering,'' ``Two-step Consensus Clustering,'' ``Biclustering,'' and ``Comparative Analysis'' to give optimal clusters as output.

\subsection{Gaze Data}
Let $\mathcal{G}$ be a gaze dataset consisting of scanpaths corresponding to (user, stimuli) tuples, which is an input to EnsembleGaze. For this study, we consider free-viewing gaze datasets that had used 2D static images as stimuli. Usually, the raw eye tracking data is preprocessed using a filter algorithm, such as Inverse Velocity Threshold (I-VT) or Dispersion Threshold Identification (I-DT), to segregate fixations from the saccades. The preprocessed gaze data is used to construct scanpaths of each participant for each stimulus. An example of a scanpath is visualized in Figure~\ref{fig:classification-grid}. A folder of stimuli images, stored as PNG or JPG files, is provided in the dataset. In cases of administering specific stimuli, additional folders are provided featuring the stimuli-IDs and the order of stimuli shown to each user. The dataset also consists of a series of fixation points, marked as $(x, y)$ screen co-ordinates, with time duration in seconds or milliseconds. Alternatively, the start and end timestamps of each fixation point may be given, which would then be used to compute the fixation duration for each tuple. User and stimulus sets usually share a many-to-many relationship, which becomes a many-to-one relationship under a between-subjects design. We determine optimal features by testing sets of features, based on both user scanpath and fixations.

\paragraph{\bf Clustering Objective-based Data Representation:} We represent the scanpath data as belonging to a specific \textit{(user, stimulus) tuple}, treated as an ordered pair. Thus, the (user, stimulus) tuple is treated as a ``key'' to each gaze-data instance, \ie~a scanpath, which is the ``value.''

 Here, we focus on the \textit{user-}, \textit{stimuli-}, and \textit{(user, stimuli)}-clustering as three different \textbf{clustering objectives}. Hereafter, we refer the entity being clustered as the \textbf{data point}. Thus, the data point is either a user, an image stimulus, or a (user, stimulus) tuple, respectively, depending on the objective. 

In the case of user- and stimuli-clustering, the covariable of the data point in its corresponding (user, stimuli) tuples is referred to as the \textbf{dual point}. 

\subsection{Image Categories}
\begin{figure*}[t]
    \centering
    \includegraphics[width=\linewidth]{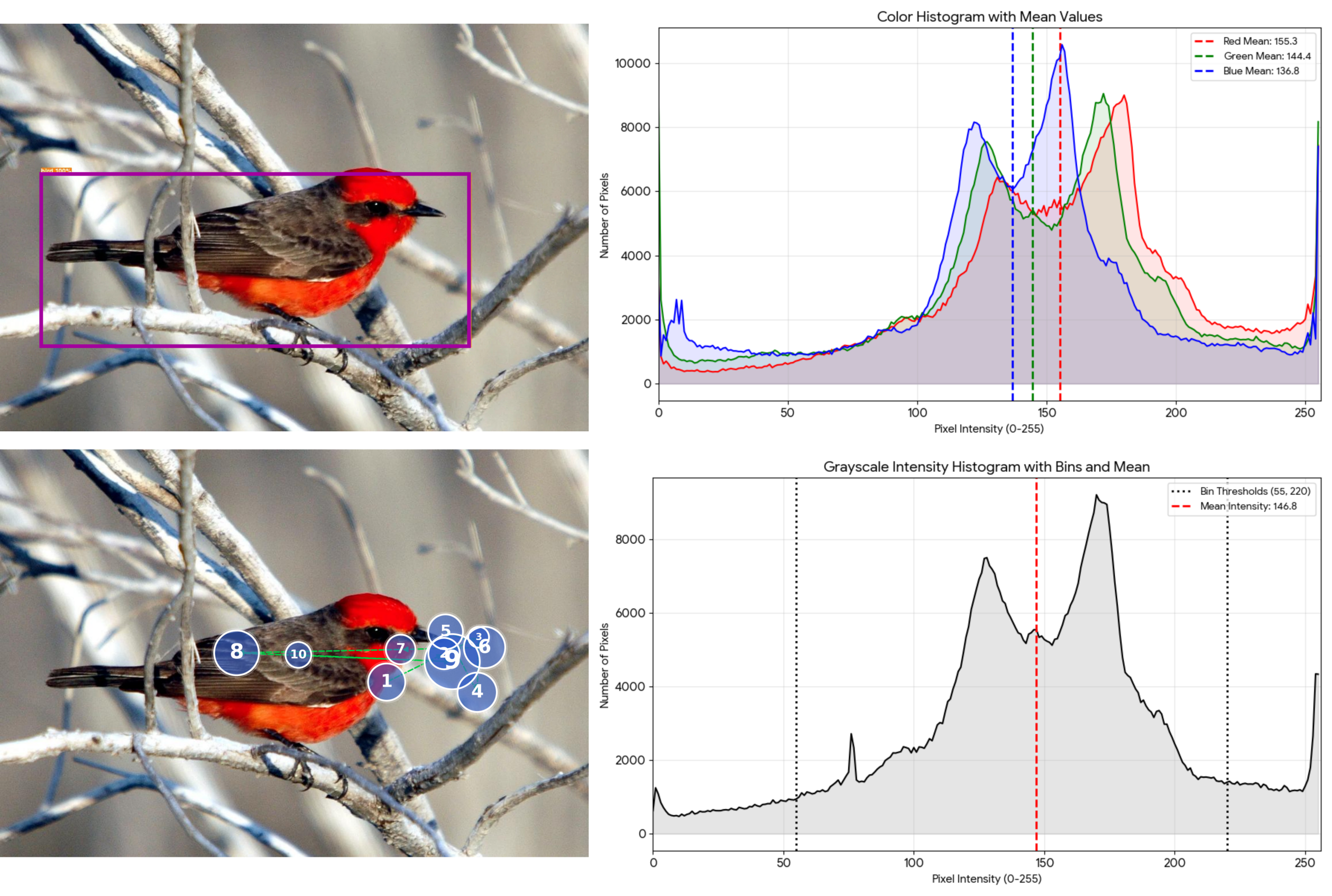}
    \caption{Image properties for a sample image from the MIT1003 dataset, with the filename `i118680869.jpeg', in which a bird object is detected using a selected deep-learning object-detection model (shown in the top-left). The scanpath of participant `ajs' is overlaid on the stimulus, where each fixation is represented as a circle numbered in order of appearance in the scanpath, with the circle radius encoding fixation duration, visualized using Eyeexplore~\cite{gnanaraj2023eyeexplore}. The plot in the top right shows a histogram of channel-wise intensity distributions and the mean values of each channel. The plot on the bottom right shows a histogram of intensities (HOI), with bins marked for low, medium, and high intensity ranges across the dataset. The bins were computed using the minimum and maximum intensity values observed across all images in the dataset.}
    \label{fig:classification-grid}
\end{figure*}

\begin{figure}[h]
    \centering
    \includegraphics[width=0.5\linewidth]{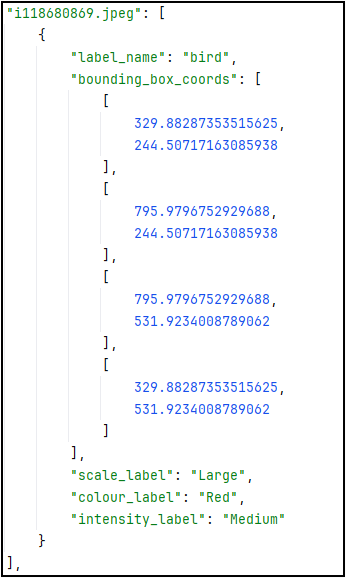}
    \caption{An example of the JSON structure representing object-level annotations, including bounding box coordinates, scale, color, and intensity labels for each detected object within an image, shown in Figure~\ref{fig:classification-grid}.}
    \label{fig:listing-2}
\end{figure}

We categorize the stimulus images based on three properties, namely, \textit{color}, \textit{intensity}, and \textit{dominant object(s) in the scene}. The objects within each image are further categorized based on their color, intensity, and scale. These color- and object-based properties are considered due to their potential influence on gaze behavior and viewing patterns. We use the data-driven categorization to compare with the stimuli-clusters computed from the gaze data.

\paragraph{\bf Color-based Categories}
Color information is extracted from each image using color histograms, that captures the distribution of pixel values across the red, green, and blue channels. The mean pixel intensity for each channel is computed to determine the predominant color of the image. The dominance of a channel is determined as the one for which there is significantly higher mean intensity than the others. The image is categorized based on its dominant channel, namely, \textit{Red}, \textit{Green}, or \textit{Blue}, which are referred to as \textbf{color-hue-based categories}. In cases where a single channel is not dominant, \eg~no channel is dominant, or the intensities are nearly equal in 2+ channels, then the image is classified as \textit{GrayScale}. The category information is stored in CSV files.

The intensity of each image is categorized based on the Histogram of Intensities (HOI), as the peak bin ($I_{peak}$) in the HOI represents the most frequent intensity value in the image. The HOI is computed on the image after grayscale conversion. Predefined intervals of frequency of intensity values are used to systematically categorize images. The intervals are based on the minimum ($I_{min}$), maximum ($I_{max}$), and mid-range ($I_{mid}$), \ie~min-max average, intensity values across all image stimuli. The \textbf{color-intensity-based categories} of images are:
\begin{itemize}
    \item \textit{Low-intensity}: with $I_{peak}<I_{LB}=0.5(I_{min}+I_{mid})$.
    \item \textit{High-intensity}: with $I_{peak}>I_{UB}=0.5(I_{max}-I_{mid})$.
    \item \textit{Medium-intensity}: with $I_{LB}\le I_{peak} \le I_{UB}$.
\end{itemize}

\paragraph{\bf Object-based Categories}
The objects present in the scene of the images and their semantics influence eye gaze. To develop the categorization, we first detect objects and their properties from the image stimuli using deep learning object-detection models. We evaluated widely used models such as GLIP~\cite{li2022grounded}, YOLO~\cite{redmon2016you}, ViT~\cite{dosovitskiy2020image}, and DETR~\cite{carion2020end}, for our requirement. Of these, YOLO produced the most consistent results and thus was selected for object detection. We detect and count instances of the objects to determine the category.

Generalizing object types usually observed in the scenes in free-viewing gaze datasets, we consider four main \textbf{object-type-based categories}, namely \textit{Wildlife}, \textit{People}, \textit{Vehicles}, and \textit{Objects}. The dominant object type in the image is the one with the highest frequency of occurrence in the scene. In case of a tie between two dominant object types, the image is assigned to both categories. 

After a high-level object-based categorization, the images are subjected to a fine-grained object-property-based categorization. For the finer-grained categorization, the object properties, namely, \textit{color}, \textit{intensity}, and \textit{scale} are considered.

The object detection model provides bounding boxes as outputs. These regions within the boxes are annotated, and the fixation metrics for the region within these boxes, \eg~fixation counts, are determined. For bounding boxes with no fixations, the metrics are set to null, and these boxes are considered \textit{non-viable}. The object properties of the viable boxes are then used for the subcategorization. The box color and intensity are computed similarly to those of the image stimuli, as explained. The YOLO-detected box diagonal is orientation-independent, and its length $d_{scale}$ provides the object scale. The \textbf{object-scale-based categories} are \textit{Small}, \textit{Medium}, and \textit{Large}, based on the percentile-based intervals of the diagonal lengths across all image stimuli. The object-scale-based categories:
\begin{itemize}
    \item \textit{Small}: where $d_{scale}<Q_{0.33}(d_{scale})$.
    \item \textit{Medium}: where $Q_{0.33}(d_{scale})\le d_{scale}\le Q_{0.66}(d_{scale})$.
    \item \textit{Large}: where $d_{scale}>Q_{0.66}(d_{scale})$.
\end{itemize}

After categorizing the images and objects based on their visual properties, a structured JSON file is created to store the details for all detected objects. For each image stimulus, its properties are stored in CSV files, and for each detected object, the JSON file contains a JSON object with attributes for the properties of the object, namely, label, the bounding box coordinates, and the corresponding visual property labels. As an example, the computations on the bounding box for a sample image from one of the datasets are illustrated in Figure~\ref{fig:classification-grid}, with its corresponding JSON structure in Figure~\ref{fig:listing-2}.

\begin{table}[t]
  \centering
  \scriptsize
  \caption{Reference list of symbols used in the paper.}
  \label{tab:symbols}
  \begin{tabular}{c|l}
    \hline
    Symbol & Description \\
    \hline
    
    $\mathcal{G}$ & Gaze dataset \\ 
    $N$ & Total number of users in $\mathcal{G}$ \\
    $S$ & Total number of image stimuli in $\mathcal{G}$ \\
    $M$ & Total number of distance metrics used in distance matrix \\
    $\mathcal{N}$ & Total number of data points based on the clustering objective, \\
    & \ie~$N$ for user-clustering, $S$ for stimuli-clustering, and \\
    & ($N\cdot S$) for (user,stimuli)-clustering \\
    \hline
    
    ($x,y$) & Screen coordinates \\
    $\mathcal{P}(i,k)$ & Scanpath of $i^{th}$ user for the $k^{th}$ image \\
    $F(i,k)$ & Total number of fixations in scanpath $\mathcal{P}(i,k)$ \\
    \hline
    
    ($\mathcal{P}_a,\mathcal{P}_b$) & A pair of scanpaths, each treated as trajectories, for comparison \\
    ($x_f,y_f$) & Spatial location of the $f^{th}$ fixation \\
    $t_f^{start} , t_f^{end}$ & Start and end time instances of the $f^{th}$ fixation \\
    $d^{(m)}(i,j)_k$ & Distance between $\mathcal{P}(i,k)$ and $\mathcal{P}(j,k)$ using $m^{th}$ metric \\
    $\mathbf{D}^{(m)}$ & Scanpath distance matrix of size $N$ using $m^{th}$ distance metric \\
    $\{d_F,d_H,d_E,$
    & \{Fr\'echet, Hausdorff, Euclidean, Dynamic Time Warping\} \\
    $,d_{DTW}\}(.,.)$ & distance measures \\
    $\mathbf{X}_d$ & Distance-based feature matrix of $\mathcal{G}$\\
    \hline
    
    $FC(i,k)$ & Total fixation count of $\mathcal{P}(i,k)$ \\
    $TTF(i,k)$ & Time-to-first-fixation in $\mathcal{P}(i,k)$ \\
    $\mathbf{FD}(i,k)$ & A vector of fixation durations for $F(i,k)$ fixations in $\mathcal{P}(i,k)$ \\
    $\mathcal{F}(i,k)$ & Set of fixation-based variables for $\mathcal{P}(i,k)$ \\
    $\mathcal{S}_D(p)$ & Set of concatenated distributions for a data point $p$ \\
    $P_D$ & Number of fixation-distribution-based features for $\mathcal{G}$ \\
    $\mathbf{f}_D(p)$ & Fixation-distribution-based feature vector of data point $p$ \\
    $\mathbf{X}_D$ & Fixation-distribution-based feature matrix of $\mathcal{G}$\\
    \hline
    
    $\mathbf{f}_T(p)(i,k)$ & Fixation-based feature vector of $\mathcal{P}(i,k)$ for tensor construction\\
    $\mathcal{X}_F$ & Fixation-based feature tensor of $\mathcal{G}$\\
    \hline

    $\mathbf{f}_{BC}$ & Feature vector of each data point used in baseline clustering \\
    $P_{BC}$ & Length of $\mathbf{f}_{BC}$  \\
    $\mathbf{X}_{BC}$ & Feature matrix of $\mathcal{G}$ for baseline clustering \\
    $R$ & Number of baseline clusterings in an ensemble \\
    $\mathcal{C}^{(r)}$ & Clustering outcome of $r^{th}$ baseline method \\
    $C^{(r)}_l$ & $l^{th}$ cluster in $\mathcal{C}^{(r)}$ \\
    $n_r$ & Number of clusters in $\mathcal{C}^{(r)}$ \\
    $\mathbf{C}^{(r)}$ & Binary co-association matrix of the $r^{th}$ baseline clustering \\
    $\mathbf{C}^{(cv)}$ & Probabilistic co-association matrix obtained from consensus \\
    & voting in an ensemble \\
    \hline
    
    $\mathbf{\Delta}^{deg}$ & Diagonal degree matrix of $\mathbf{C}^{(cv)}$\\
    $\mathbf{L}_{sym}$ & Symmetric normalized Laplacian matrix of $\mathbf{C}^{(cv)}$ \\
    $\mathbf{\Gamma}$ & Dissimilarity matrix computed as the complement of $\mathbf{C}^{(cv)}$ \\
    $\mathbf{u}$ & Vectorization of the strictly upper triangular matrix of \\
    & $\mathbf{\Gamma}$ in the row-major order \\
    $\mathcal{Z}$ & Dendrogram of hierarchical agglomerative clustering of $\mathbf{\Gamma}$ \\
    $\mathcal{C}^\ast$ & Final consensus clustering of optimal $k$ clusters of $\mathcal{G}$ \\
    \hline

    $\mathbf f_{SC}(i)^{S}$ & Fixation-based feature vector of $\mathcal{P}(i,\cdot)$ for \\
    & stimulus-guided subspace clustering \\
    $\mathbf f_{SC}(k)^{U}$ & Fixation-based feature vector of $\mathcal{P}(\cdot,k)$ for \\
    & user-guided subspace clustering \\
    $n_{PC}$ & Number of primary clusters in subspace clustering \\
    $\mathbf X_{flat}^{U}$ & Feature matrix from mode-1 unfolding of $\mathcal X_F$, \\
    & \ie~user-guided \\
    $\mathbf X_{flat}^{S}$ & Feature matrix from mode-2 unfolding of $\mathcal X_F$, \\
    & \ie~stimulus-guided \\
    \hline

  \end{tabular}
\end{table}

\subsection{Feature Engineering} \label{sec:feature_selection}
We extract specific feature sets from the gaze dataset, $\mathcal{G}$, which are inputs to the unsupervised learning models. Feature engineering, here, involves transforming the scanpath data into feature sets that can be aggregated for each data point. For instance, to study the user behavior patterns in gaze data, user-clustering is implemented using the users' aggregated feature set. On the other hand, stimuli-clustering is needed for studying saliency using aggregated feature sets of image stimuli, thus revealing patterns within the chosen stimulus set. While the optimal choice of feature set is guided by the clustering objective, our work shows that the rudimentary features that are available across most gaze datasets are sufficiently rich to mine interesting patterns.

We use two \textbf{feature types}: \textit{Scanpath-Distance-} and \textit{Fixation-}based features. Here, the distance-based features are used exclusively in user-clustering in the form of a \textit{feature matrix}. The distribution-based features are more versatile and are used in both user- and stimuli-clustering as a \textit{feature matrix} representation, and in (user, stimuli)-clustering as a \textit{feature tensor} representation. Table~\ref{tab:symbols} gives a reference list of all the symbols used hereafter.

\paragraph{\bf Scanpath-Distance-based Features}
Here, the scanpath of each user is a \textit{spatial trajectory} data. A distance matrix is computed using geometric distance between scanpaths of any two users. We use various distance measures, such as Fr\'echet distance, Hausdorff distance, Euclidean scanpath distance, dynamic time warping (DTW), and string-based time-delay-embedding distance (SBTDED). $D^{(m)}$ is the distance matrix computed using the $m^{th}$ distance measure, where $m=\{1,\ldots,M\}$ for $M$ selected measures. We first compute the distance between scanpaths of each pair of users $(i, j)$ for each image, and $D^{(m)}_{ij}$ is the summation of distances measured for each pair of users $(i, j)$ across all images. The matrix $D$ based on distance of scanpaths is now included in the feature set.

For the $i^{th}$ user observing the $k^{th}$ image stimulus in a free-viewing gaze dataset of $N$ users and $S$ image stimuli, the scanpath is represented as an ordered sequence of fixations:
\[
\mathbf{\mathcal{P}}(i,k) = \{(x_f, y_f, t_f^{\text{start}}, t_f^{\text{end}})\}_{f=1}^{F_{ik}},
\]
where \( (x_f, y_f) \) denotes the spatial location of the $f^{th}$ fixation, and \( t_f^{\text{start}} \) and \( t_f^{\text{end}} \) denote the temporal onset and offset of the fixation, respectively. Here, \(F_{ik}\) represents the total number of fixations produced by the (user, stimulus) tuple, ($i,k$).

For every pair of users \((i,j)\), a scanpath distance is computed independently for each image \(k\) using multiple trajectory similarity measures. 
\begin{equation}
    d^{(m)}(i,j)_k=d^{(m)}\left(\mathbf{\mathcal{P}}(i,k),\mathbf{\mathcal{P}}(j,k)\right), \nonumber
\end{equation}
where \(m^{th}\) distance measure is used. Each element in the distance matrix is a pairwise user dissimilarity summed across all $S$ stimuli, given as:
\begin{equation}
    \mathbf{D}^{(m)}_{ij}=\sum_{k=1}^S d_j^{(m)}(i,j)_k, \nonumber
\end{equation}
Thus, the resulting matrix $\mathbf{D}^{(m)} \in \mathbb{R}^{(N \times N)}$ represents the pairwise scanpath dissimilarity between all participant pairs using the $m^{th}$ distance measure. The distance-based feature matrix of the dataset is formally represented as $\mathbf{X}_d(\mathcal{G})=\mathbf{D}^{(m)}$.

We perform a systematic analysis of multiple trajectory-based similarity measures, that capture both spatial and temporal characteristics of gaze behavior. Given two scanpaths, $\mathbf{\mathcal{P}}_a$ and $\mathbf{\mathcal{P}}_b$:
\begin{itemize}
    \item \textbf{Euclidean Scanpath Distance}:  Measures point-wise spatial deviation between corresponding fixation coordinates:
    \begin{equation}
        d_E(\mathbf{\mathcal{P}}_a,\mathbf{\mathcal{P}}_b)=\sqrt{\sum_{k=1}^{L}\|s_k^{(a)} - s_k^{(b)}\|^2}, \nonumber
    \end{equation}where \(L\) denotes the aligned scanpath length.
    \item \textbf{Fr\'echet Distance}:  Measures the spatial similarity between two trajectories while preserving the sequential ordering of points. The discrete Fr\'echet distance is defined as:
    \begin{equation}
        d_F(\mathbf{\mathcal{P}}_a,\mathbf{\mathcal{P}}_b)=\inf_{\alpha,\beta}\;\max_t\|\mathbf{\mathcal{P}}_a(\alpha(t)) - \mathbf{\mathcal{P}}_b(\beta(t))\|, \nonumber
    \end{equation}
    where \(\alpha(t)\) and \(\beta(t)\) define monotonic traversals along the two trajectories.
    \item \textbf{Hausdorff Distance}:  Captures the maximum deviation between two sets of fixation points:
    \begin{equation}
        \begin{split}
        d_H(\mathbf{\mathcal{P}}_a,\mathbf{\mathcal{P}}_b)=
        \max\{& \sup_{x \in \mathbf{\mathcal{P}}_a}\inf_{y \in \mathbf{\mathcal{P}}_b}\|x-y\|, \\
        & \sup_{y \in \mathbf{\mathcal{P}}_b}\inf_{x \in \mathbf{\mathcal{P}}_a}\|y-x\|\}. \nonumber
        \end{split}   
    \end{equation}
    \item \textbf{Dynamic Time Warping (DTW)}:  Accounts for temporal misalignment between scanpaths by optimally warping fixation sequences:
    \begin{equation}
        d_{DTW}(\mathbf{\mathcal{P}}_a,\mathbf{\mathcal{P}}_b)=\min_{\pi}\sum_{(u,v)\in\pi}\|s_u^{(a)} - s_v^{(b)}\|, \nonumber
    \end{equation}
    where \(\pi\) denotes the optimal warping path.
    \item \textbf{String-Based Time Delay Embedding (SBTDED)}:  Encodes scanpaths as symbolic temporal sequences to capture higher-order temporal dependencies and sequential viewing behavior. Distance is computed over the embedded symbolic representations of gaze transitions.
\end{itemize}
Collectively, these selected metrics quantify gaze behavior by capturing spatial correspondence, temporal alignment, trajectory continuity, and sequential fixation dynamics across users.

\paragraph{\bf Fixation-based Features}
Fixations indicate user-attention on particular areas in a stimulus. Since our objective is to group data points based on their cognitive process and attention allocation, as stated in Section~\ref{sec:lit_feature_engineering}, we use fixations to engineer features for clustering. We compute the fixation-based distributions for each scanpath corresponding to a (user, stimulus) tuple. The three \textbf{fixation-based variables} are \textit{total fixation counts} ($FC$), \textit{time-to-first-fixation} ($TTF$), and \textit{fixation durations} ($\mathbf{FD}$). Figure~\ref{fig:classification-grid} shows a scanpath with $FC$=10, and these fixations are visualized as circles with radius proportional to the fixation durations. These duration values form a discrete distribution $\mathbf {FD}$ of length 10. 

\noindent{\it\bf Feature Matrix:} We compute the \textit{frequency distribution} of each variable, and subsequently, the five \textbf{descriptive statistical measures} as distribution-based features for each tuple. These statistical measures are \textit{total}, \textit{mean}, \textit{median}, \textit{standard deviation}, and \textit{variance} of the tuple-specific distributions. Based on the clustering objective, we concatenate the values at the data-point-level to construct the frequency distributions, \ie~$\bigoplus$($i,\cdot$) for user- and $\bigoplus$($\cdot,k$) for stimuli-clustering, respectively. For each of the available concatenated distributions, we apply the statistical descriptors, represented as an operator set,\\
\centerline{$\Theta=\bigg\{\sum, \mu, Median, \sigma, \sigma^2\bigg\}$.}\\

From each scanpath \(\mathcal{P}(i,k)\) with $F(i,k)$ fixations, a set of fixation-based gaze metrics is extracted. Let
\begin{equation}
\begin{split}
\mathcal{F}(i,k) = \{&FC(i,k), TTF(i,k), \mathbf{FD}(i,k) \mid \\
&FC(i,k),TTF(i,k)\in\mathbb{Z}^+, \\
&\mathbf{FD}(i,k)\in(\mathbb{Z}^+)^{F(i,k)}\} \nonumber
\end{split}
\end{equation}
denote the set of fixation-based variables.

For a data point, the set of concatenated distributions for user-clustering are:
\begin{equation}
\begin{split}
\mathcal{S}_D(i)&=\mathcal{F}(i,.) \\
&=\bigg\{\bigoplus\limits_{k=1}^S FC(i,k), \bigoplus\limits_{k=1}^S TTF(i,k), \bigoplus\limits_{k=1}^S \mathbf{FD}(i,k)\bigg\},\nonumber
\end{split}
\end{equation}
and for stimuli-clustering are:
\begin{equation}
\begin{split}
\mathcal{S}_D(k)&=\mathcal{F}(.,k) \\
&= \bigg\{\bigoplus\limits_{i=1}^N FC(i,k), \bigoplus\limits_{i=1}^N TTF(i,k), \bigoplus\limits_{i=1}^N \mathbf{FD}(i,k)\bigg\}. \nonumber
\end{split}
\end{equation}
It must be noted that while the sizes of the support of the distributions of $\bigoplus FC$ and $\bigoplus TTF$ are the number of scanpaths used for concatenation, but that of $\bigoplus \mathbf{FD}$ is the sum of fixations across all the scanpaths in the concatenation.

Thus, the generalized fixation-distribution-based feature vector $\mathbf{f}_D$ for a data point $p$ is:
\begin{equation}
\begin{split}
\mathbf{f}_D(p)=\{& o_a(v_b)\mid \\
& o_a\in \Theta,v_b\in\mathcal{S}_D(p), 1\le a\le 5, 1\le b\le N_F\}, \nonumber
\end{split}
\end{equation}
where $v_b$ is the distribution with regard to a variable in the set $\mathcal{S}_D(p)$, and $N_F$ is the size of $\mathcal{S}_D(p)$.

A feature matrix is constructed by using the feature vectors of all data points as rows. The fixation-distribution-based feature matrix of the dataset is:
\[
\mathbf{X}_D(\mathcal{G})=[\mathbf{f}_D (p_1),\ldots,\mathbf{f}_D (p_{\mathcal{N}})]^T,
\]
where \(\mathcal{N}\) denotes the number of data points based on the clustering objective, \ie~$\mathcal{N}=N$ for user-clustering, and $\mathcal{N}=S$ for stimuli-clustering.

\noindent{\it\bf Feature Tensor:} For (user, stimuli)-clustering, we construct a fixation-based feature vector $\mathbf{f}_T$ using selected fixation-based measures of the (user, stimulus) already described in the distribution-based features. The feature vector, $\mathbf{f}_T\in(\mathbb{R}^+)^5$, for scanpath $\mathcal{P}(i,k)$ is:
\begin{equation}
\begin{split}
\mathbf{f}_T(i,k)=\bigg[&\sum(\mathbf{FD}(i,k)),\mu(\mathbf{FD}(i,k)),\sigma^2(\mathbf{FD}(i,k)),\\
&FC(i,k),TTF(i,k)\bigg] \nonumber
\end{split}
\end{equation}

Thus, putting together the vectors corresponding to all the scanpaths in the dataset, $\mathcal{G}$, we get the fixation-based feature tensor, $\mathcal{X}_F\in\mathbb{R}^{(N\times S\times 5)}$, is: 
\begin{equation}
\mathcal{X}_F(\mathcal{G}) = (\mathbf{f}_T(i,k))_{N\times S} \text{, where }  \mathbf{f}_T\in(\mathbb{R}^+)^5\nonumber
\end{equation}

The usage of the tensor using projection for (user,stimuli)-clustering is explained in Section~\ref{sec:hdc}.

\subsection{Consensus Clustering} \label{sec:consensus}
In the absence of ground truth for clustering in gaze data, we address two goals, based on the selected clustering objective:
\begin{enumerate}
\item To estimate the optimal number of distinct, non-overlapping clusters required to partition data points.
\item To determine the optimal clustering of the data points.
\end{enumerate}
Since EDA of gaze data using novel clustering objectives is underexplored, we focus on hard clustering as a first cut. As with all real-world datasets, soft/fuzzy clustering needs an investigation, which is considered as future work.

We use the novel strategies of consensus clustering and co-clustering free-viewing gaze data in EnsembleGaze. The rationale behind using consensus clustering is that, in baseline unsupervised models, the clustering performance is sensitive to initialization conditions, noise and outliers, and has an inherent bias toward specific cluster geometries. Since latent semantic structures within the dataset are revealed non-uniformly across different clustering algorithms differently, we propose unsupervised ensemble learning in EnsembleGaze to obtain a more robust and stable partitioning.

For consensus clustering, we first identify the baseline unsupervised models to be used in the ensemble, and then identify an appropriate consensus voting and meta-learning model for the final optimal clustering.

\paragraph{\bf Baseline Clusterings}
The baseline clustering models provide both the optimal number of clusters, $k$, and the partitionings to be included in the ensemble. The optimal $k$ is determined through a combination of performance metrics, such as Silhouette index, Davies Bouldin index, and Calinski Harabasz index. The results from these metrics for each feature set are manually examined to determine the optimal $k$, which is then initialized for any clustering experiment needing $k$ as an input. 

We choose three widely used clustering algorithms in the ensemble, namely $k$-means (apt for large sample sizes), hierarchical agglomerative (apt for medium sample sizes) and spectral (apt for smaller sample sizes, where non-convex clusters are present). We use a \textit{heterogeneous ensemble}~\cite{iam2011link} to provide a more robust clustering that satisfies a diverse set of clustering criteria. This is a robust approach for unsupervised clustering in the absence of ground truth.

\paragraph{\bf Ensemble Clustering and Consensus Voting}\label{method:consensus_clustering}
In this study, we employ co-association matrix-based ensemble clustering, which uses an Evidence Accumulation Clustering (EAC)~\cite{fred2005combining}. The consensus voting is implemented by averaging the binary co-association matrices of the baseline clusterings of the data points, to give the \textit{probabilistic} co-association matrix. 

Given feature vector $\mathbf{f}_{BC}(p)\in\mathbb{R}^{P_{BC}}$ of a data point $p$, for $\mathcal{N}$ data points, the feature matrix for baseline clustering is:
\[
\mathbf{X}_{BC}(\mathcal{G})=[\mathbf{f}_{BC} (p_1),\ldots,\mathbf{f}_{BC} (p_{\mathcal{N}})]^T,
\]

Applying $R$ baseline methods independently to $\mathbf{X}_{BC}$ gives a set of clusterings:
\[
\mathcal{C}^{(1)}, \mathcal{C}^{(2)}, \dots, \mathcal{C}^{(R)}.
\]

The $r^{th}$ baseline clustering method produces a partitioning with $n_r$ clusters:
\[
\mathcal{C}^{(r)} = \left\{C^{(r)}_1,\dots,C^{(r)}_{n_r}\right\}.
\]

Each baseline clustering method produces a partitioning of the data points in $\mathcal{G}$, resulting in a \textit{many-to-one mapping} of data points to cluster IDs, \ie~each data point gets a cluster ID label. Such a partitioning is also represented as a binary co-association matrix whose elements are pairwise cluster membership relationships between data points. For the $r^{th}$ baseline clustering, the co-association matrix, $\mathbf{C}^{(r)}\in\{0,1\}^{\mathcal{N}\times\mathcal{N}}$, is defined as:
\[
\mathbf{C}^{(r)}_{ij} =
\begin{cases}
1, & \text{if data points } i \text{ and } j \text{ belong to the same cluster}, \\
0, & \text{otherwise}.
\end{cases}
\]

The final ensemble co-association matrix $\mathbf{C}^{(cv)}$ is a probability matrix, which is obtained by a consensus voting mechanism, \ie~averaging across all clustering methods:
\[
\mathbf{C}^{(cv)} = \frac{1}{R}\sum_{r=1}^{R} C^{(r)}.
\]

\paragraph{\bf Meta-Learning for Final Clustering}
We use Hierarchical Agglomerative Clustering (HAC)~\cite{sokal1958statistical,fred2005combining} as a meta-learning clustering algorithm on the final ensemble co-association matrix to obtain the optimal clustering. 

In the HAC implementation, a dissimilarity matrix corresponding to the final ensemble co-association matrix is used. The dissimilarity matrix is:\\
\centerline{$\mathbf{\Gamma} = \mathbf{1} - \mathbf{C}^{(cv)}$},\\
where \( \mathbf{1} \) denotes an all-ones matrix. $\mathbf\Gamma$ is also a probability matrix, similar to $\mathbf{C}^{(cv)}$ and is the complement of $\mathbf{C}^{(cv)}$. Since $\mathbf{\Gamma}$ is symmetric with 0s in the diagonal, we use the vectorization of the \textit{strictly upper triangular matrix} of $\mathbf{\Gamma}$ in the row-major order, \ie~the upper triangular elements without the diagonal elements. The vectorization gives the condensed vector $\mathbf{u}$ given by:
\[
\mathbf{u} = \text{vec}_{\triangle}(\mathbf{\Gamma}),
\]
where \( \text{vec}_{\triangle}(\cdot) \) denotes the vectorization of the strictly upper triangular elements of \( \mathbf{\Gamma} \), excluding the diagonal. An HAC procedure with average linkage is applied to $\mathbf{u}$ to obtain a dendogram $\mathcal{Z}$, given as:
\[
\mathcal{Z} = \text{HClust}(\mathbf{u}, \text{linkage}=\text{average}).
\]

The HAC starts by treating every data point as a \textit{singleton} cluster. Then, at each merge step, the inter-cluster distance between clusters \( C_a \) and \( C_b \) is defined as:
\[
\delta(C_a, C_b) = \frac{1}{|C_a||C_b|} \sum\limits_{i \in C_a} \sum\limits_{j \in C_b} \mathbf{\Gamma}_{ij}.
\]

Then, using the optimal $k$ determined by the baseline clustering models, $\mathcal{Z}$ is cut to obtain $k$ optimal clusters, \ie~optimal clustering $\mathcal{C}^\ast$ of data points $\{p_1,\ldots,p_\mathcal{N}\}$ with cluster IDs $c(\cdot)$, using a maximum-cluster criterion:
\begin{equation}
\begin{split}
  \mathcal{C}^{\ast} & = \text{Cut}(\mathcal{Z}, k) \\
  & = \{c(p_1),\ldots,c(p_\mathcal{N})\} \text{, where } 1\le c(\cdot) \le k. \nonumber
\end{split}
\end{equation}

\subsection{High-Dimensional Clustering} \label{sec:hdc}
The high-dimensional clustering step enables us to consider the (user, stimulus)-identity of the scanpaths in the gaze data for clustering, as opposed to using the user or stimuli information exclusively, as done in one-way clustering (Section~\ref{sec:consensus}). Consensus subspace clustering and biclustering are the two proposed high-dimensional clustering approaches for gaze data. The consensus subspace clustering uses a similar methodology as explained further in Section~\ref{sec:consensus}, and a sequential, nested, and top-down strategy. On the other hand, biclustering, is a strictly simultaneous clustering method.

\paragraph{Consensus Subspace Clustering}
Our strategy for subspace clustering is to compute \textit{primary clustering} using one of the two dimensions, \ie~users or image stimuli, as the data point, and then compute clusters within the primary clusters using the dual point as the clustering objective. Properties of our approach:
\begin{itemize}
\item This method is \textit{sequential} in performing a two-step clustering, where the clustering across both dimensions are highly decoupled.
\item It resembles a hierarchical divisive clustering technique, which makes it \textit{top-down}.
\item The final clusters are \textit{nested} within the primary clusters.
\item This method is a \textit{subspace clustering} strategy, as the primary clusters are assumed to possess their own unique subspace characterization.
\end{itemize}

For free-viewing gaze data, we use two different sequences of dimensions for implementation. The sequences and their corresponding feature vectors computed from the feature vectors $\mathbf f_T$ used to construct feature tensor $\mathcal X_F$ are:
\begin{itemize}
\item \textit{User-guided} sequence: $\mathbf f_{SC}(i)^{U} = \frac{1}{S}\sum\limits_{k=1}^S \mathbf f_T(i,k)$.
\item \textit{Stimulus-guided} sequence: $\mathbf f_{SC}(k)^{S} = \frac{1}{N}\sum\limits_{i=1}^N \mathbf f_T(i,k)$.
\end{itemize}
The feature vectors corresponding to the selected sequence are used to compute the primary clusters of the concerned data points, \ie~users and stimuli for user- and stimulus-guided sequences, respectively.

In the second step of subspace clustering, user- and stimulus-guided sequences perform stimuli- and user-clusterings, respectively. Here, the corresponding feature vectors are computed by averaging within the primary clusters. Thus, for primary clusters $G_1,\ldots,G_{n_{PC}}$, the feature vectors used for clustering within the $j^{th}$ primary cluster in the second step of the sequences are:
\begin{itemize}
\item \textit{User-guided} sequence: $\mathbf f_{SC}(k)^{S} = \frac{1}{|G_j|}\sum\limits_{i\in G_j} \mathbf f_T(i,k)$.
\item \textit{Stimulus-guided} sequence: $\mathbf f_{SC}(i)^{U} = \frac{1}{|G_j|}\sum\limits_{k\in G_j} \mathbf f_T(i,k)$.
\end{itemize}

We propose consensus clustering using heterogeneous ensembles in each step of subspace clustering sequences, similar to the one-way clustering (Section~\ref{sec:consensus}). For each ensemble, we use the same set of selected baseline clustering mehods, namely, $k$-means, hierarchical agglomerative, and spectral methods. Overall, each clustering step in subspace clustering is implemented using consensus clustering, involving baseline clustering, co-association matrix-based consensus voting, and HAC-based meta learning. Thus, we refer to this method as the \textit{consensus subspace clustering} (CSC).

\paragraph{Biclustering} 
Biclustering is the process of grouping (user, stimuli) tuples, \ie~grouping \textit{submatrices} in a data matrix of (user $\times$ stimuli). Specifically, biclusters reveal patterns across both the rows and the columns of the fixation-based feature tensor, $\mathcal X_F(\mathcal G)$, whose construction is explained in Section~\ref{sec:feature_selection}. Spectral biclustering method uses $\mathcal X_F$ through matrix unfolding. 

\begin{definition}
    \textbf{Matrix Unfolding}, also referred to as tensor matrixicization or flattening, is the operation that reshapes a 3D tensor into a matrix by concatenating slices along a specific dimension, referred to as a mode.
\end{definition}

In our case of $\mathcal X_F\in \mathbb (R^+)^{N\times S\times 5}$, mode-1 unfolding along the rows leads to slicing the rows, \ie~the users. Similarly, mode-2 unfolding slices along the columns, \ie~the image stimuli. Thus, we refer to mode-1 and mode-2 unfolding as \textit{user-guided} and \textit{stimulus-guided} unfolding of $\mathcal X_F$. The unfolded feature matrices, using Kolda ordering, are:
\begin{itemize}
\item User-guided unfolding: $\mathbf X_{flat}^{U}\in\mathbb (R^+)^{N\times 5S}$ is given as:
  \[
  \mathbf X_{flat}^{U}(i,c) = \mathcal X_F(i,k,f) \text{, where } c=k+(f-1)S.
  \]
\item Stimulus-guided unfolding: $\mathbf X_{flat}^{S}\in\mathbb (R^+)^{S\times5N}$ is constructed as:
  \[
  \mathbf X_{flat}^{S}(k,c) = \mathcal X_F(i,k,f) \text{, where } c=i+(f-1).
  \]
\end{itemize}

Biclustering these feature matrices give row clusters that co-activate on column clusters. For instance, $\mathbf X_{flat}^{U}$ has user-clustering along rows, and stimuli-clustering along columns; and vice versa for $\mathbf X_{flat}^{S}$. This systematic analysis of biclustering gives insights into the dominance of user behavior or image properties in gaze data generation. 

\begin{definition}
  A \textbf{bicluster} is the Cartesian product of a row group and a column group. In spectral biclustering, for a tensor element $\mathcal X_F(i,k,\cdot)$ in the form of a vector, with $c_{row}[i]$ as the row cluster label vector of the $i^{th}$ row and $c_{col}[k]$ as the column cluster label vector of the $k^{th}$ column, the cluster labels $c_{row}$ and $c_{col}$ are obtained by applying $k$-means separately on the spectral embeddings. In this scenario, if $R_r = \{i: c_{row}[i] =r\}$ and $C_c = \{k: c_{col}[k] = c\}$ are the sets of rows and columns with row- and column-cluster labels as $r$ and $c$, respectively, then the $\text{Bicluster}(r, c) = R_r \times C_c$, \ie~a Cartesian product of the row and column index sets.
\end{definition}

Let $k_{row}$ and $k_{col}$ be the optimal number of clusters of rows and columns, respectively, obtained by applying $k$-means separately to the spectral embeddings from the singular value decomposition of $\mathbf X_{flat}^{U}$ and $\mathbf X_{flat}^{S}$, respectively. Then, the total number of biclusters for the gaze dataset is $k_{row}\times k_{col}$.

In our work, we assume that the free-viewing gaze data has biclustering patterns. To implement spectral biclustering method, we predetermine $k_{row}$ and $k_{col}$. This is done by computing the silhouette scores for possible $k$-value based on the rows of $\mathbf X_{flat}^{U}$ and $\mathbf X_{flat}^{S}$, respectively. This $(k_{row}, k_{col})$ value is used to initialize the hyperparameters of the spectral biclustering method. The algorithm first normalizes the data matrix using log normalization, and then applies spectral decomposition (Singular Value Decomposition) to compute eigenvectors of the normalized matrix, which are used to project the data into a lower-dimensional subspace. Finally, $k$-means clustering algorithm is applied to these eigenvectors to determine the optimal row and column partitions, effectively grouping related entities together. 

Each bicluster is characterized by its mean activation in the standardized matrix $\mu_{(r, c)}$, \ie~the mean z-score activation of the bicluster ($r, c$). A bicluster with  $\mu_{(r, c)} > 0$ indicates that this group of rows shows above-average values of the features indexed by $C_c$, while  $\mu_{(r, c)} < 0$ indicates below-average values. The canonical diagnostic for a successful biclustering is the ``checkerboard pattern'' in the reordered mean activation matrix. 

\subsection{Comparative Analysis}
We use the standard classification validation metrics to evaluate the goodness-of-fit of each partitioning strategy. Specifically, accuracy, precision, recall, F1-score, Overall accuracy, and $\kappa$ coefficient are computed by treating the consensus clustering output as the reference/baseline clustering for comparative analysis. For $TP$, $TN$, $FP$, and $FN$ as true positives, true negatives, false positives, and false negatives, respectively; and  $p_o$ and $p_e$ are the observed and the expected proportions of agreement, respectively:

\begin{eqnarray}
\text{Accuracy }& A &= \frac{TP + TN}{TP + TN + FP + FN} \nonumber \\
\text{Precision }& P &= \frac{TP}{TP + FP} \nonumber \\
\text{Recall }& R &= \frac{TP}{TP + FN} \nonumber \\
\text{F1 Score }& F_1 &= 2\times\frac{P\cdot R}{P+R} \nonumber \\
\text{Overall Accuracy }& OA &= \frac{1}{N}\sum\limits_{i=1}^k TP_i \nonumber \\
\text{Kappa Coefficient }& \kappa &= \frac{P_o - P_e}{1 - P_e} \nonumber
\end{eqnarray}

In unsupervised learning settings, cluster distributions are often imbalanced. To address this, we additionally compute the Matthews Correlation Coefficient ($MCC$), which provides a robust measure of classification quality even under class imbalance. The MCC is defined as:
\begin{eqnarray}
MCC &= \frac{c.s-\sum\limits_{k=1}^K p_k.t_k}{\sqrt{\big(s^2-\sum\limits_{k=1}^K p_k^2\big).\big(s^2-\sum\limits_{k=1}^K t_k^2\big)}}, \nonumber
\end{eqnarray}
\begin{eqnarray}
\text{where}& c &=\sum\limits_{k=1}^K TP_k \text{, total number of correct predictions}\nonumber \\
& s &= N \text{, total support} \nonumber \\
& t_k &= TP_k+FN_k  \nonumber \\
& p_k &= TP_k+FP_k  \nonumber
\end{eqnarray}
Here, $t_k$ and $p_k$ represent the total numbers of actual instances and predicted instances of the $k^{th}$ class, respectively.

In many gaze datasets, $S \gg N$, resulting in larger and denser stimuli-clusters. This imbalance leads to visual clutter, that makes direct visualization and interpretation challenging. To address this, we perform preliminary EDA of the (user, stimuli)-clusters using \textit{community detection} and \textit{clique analysis}. These methods facilitate the identification of cohesive substructures within clusters and provide an interpretable baseline for comparing clustering performance using the aforementioned metrics.

The interpretation of spectral biclustering results involves identifying and evaluating coherent submatrices (biclusters), where subsets of rows and columns exhibit locally consistent patterns, often described as a \textit{checkerboard structure}. To support qualitative analysis, the following visualization techniques are used:
\begin{itemize}
    \item Alluvial plots for comparing cluster memberships across selected clusterings. 
    \item Stacked bar charts for inter-cluster comparison of compositional data from a single clustering.
    \item Matrix visualizations of co-association matrices to visualize grouping patterns along the diagonal, after reordering (or seriation).
    \item Reordered activation heatmaps and partition-wise feature composition charts. 
\end{itemize}

In addition to visual analysis, we assess clustering quality using Adjusted Rand Index ($ARI$) and Normalized Mutual Information ($NMI$), respectively, which are complementary quantitative measures of agreement between clustering schemes. While $ARI$ measures pairwise agreement corrected for chance, $NMI$ measures overall information overlap between two clustering schemes. 

\begin{equation} 
ARI = \frac{RI - E[RI]}{Max(RI) - E[RI]}, \nonumber
\end{equation}
where \(RI\) is the Rand Index, \(E[RI]\) is the expected value of the Rand Index, and \(Max(RI)\) is the maximum possible Rand Index value. Mathematically, Rand Index is the same as Accuracy for clustering schemes. 

\begin{equation} 
NMI(U,V) = \frac{2\, I(U;V)}{H(U) + H(V)},
\end{equation}
where $U$ and $V$ are two clustering schemes, and the mutual information $I(U;V)$ and entropy $H(\cdot)$ are given by:

\begin{equation}  
I(U;V) = \sum\limits_{i=1}^{|U|} \sum\limits_{j=1}^{|V|} P(i,j)
\log \left(\frac{P(i,j)}{P(i)P(j)}\right)
\end{equation}

\begin{equation} 
H(U) = -\sum_{i=1}^{|U|} P(i)\log P(i)
\end{equation}

$ARI$ can range from -1 to 1, where -1 indicates discordant/dissimilar clusters, and 1 indicates perfectly similar clusters. $NMI$ ranges from 0 to 1 where 0 indicates no mutual information shared between the two clustering schemes, and 1 indicating perfect correlation between the two schemes. Prior to computation of metrics, Hungarian label alignment~\cite{munkres1957algorithms} is done with respect to the baseline scheme, in order to address the label permutation ambiguity. 

\section{Experiments and Results}\label{sec:results}
We ran experiments on EnsembleGaze with two widely used free-viewing gaze datasets. We discuss the clustering outcomes of EnsembleGaze for each dataset as a case study.

\subsection{Datasets} \label{sec:datasets}
We use the MIT1003~\cite{judd2009learning} and the EMOd (Emotional Attention Dataset)~\cite{fan2018emotional} datasets, where the former is a natural-scene viewing dataset, and the latter is a dataset of emotion-eliciting images. Both datasets provide multivariate multi-dimensional gaze data, with fixations, saccades, saliency maps, attention map, and temporal information of gaze. While the raw gaze data is not provided, both datasets are sufficiently rich to be mined for patterns using unsupervised ensemble learning. Here, we also provide the observations on the image categories for both datasets.

\begin{table}[b]
\centering
\caption{Class distribution of images for color-based categories in the datasets.}
\label{tab:color_classes}
\begin{tabular}{c|ccc|ccc}
\hline
Dataset & \multicolumn{3}{c|}{Color-hue-based}
& \multicolumn{3}{c}{Color-intensity-based} \\\cline{2-7}
(\# Images) & Red & Green & Blue & Low & Medium & High \\\hline
\bf MIT1003 (1,003) & \bf 563 & 207 & 233 & 101 & \bf 888 & 14 \\
\bf EMOd (698) & 174 & 135 & \bf 389 & 82 & \bf 578 & 38 \\
\hline
\end{tabular}
\end{table}

\begin{table*}[t]
\centering
\caption{Distribution of YOLOv5 detected object categories in the datasets, listing specific labels, total object counts, and image counts per category present in the datasets.}
\label{tab:object_distribution}
\begin{tabular}{m{4em} | m{25em}| m{1cm} | m{1cm} | m{1cm} | m{1cm} } 
\hline
&& \multicolumn{2}{c|}{\bf MIT1003} & \multicolumn{2}{c}{\bf EMOd} \\ \cline{3-6}
\bf Category & \bf YOLO Labels & \textbf{Object Count} & \textbf{Image Count} & \textbf{Object Count} & \textbf{Image Count}\\ \hline
  Objects & 'baseball bat', 'spoon', 'orange', 'vase', 'fork', 'toothbrush', 'bed', 'suitcase','frisbee', 'keyboard', 'kite', 'wine glass', 'hot dog', 'donut', 'umbrella', 'sink', 'bottle', 'remote', 'tie', 'couch', 'skateboard', 'oven', 'carrot', 'handbag', 'cell phone', 'parking meter', 'teddy bear', 'skis', 'cake', 'cup', 'traffic light', 'fire hydrant', 'book', 'microwave', 'pizza', 'potted plant', 'mouse', 'backpack', 'knife', 'sandwich', 'scissors', 'broccoli', 'laptop', 'apple', 'snowboard', 'stop sign', 'toilet', 'bench', 'dining table', 'sports ball', 'clock', 'surfboard', 'chair', 'bowl', 'banana', 'tv' & 1,046 & 360 & 564 & 227  \\ 
  \hline
  People & 'person' & \bf 1,790 & \bf 444 & \bf 1,340 & \bf 282 \\ 
  \hline
  Vehicles & 'car', 'bicycle', 'motorcycle', 'airplane', 'bus', 'train', 'truck', 'boat' & 762 & 230 & 297 & 105 \\ 
  \hline
  Wildlife & 'cat', 'dog', 'bird', 'horse', 'sheep', 'cow', 'elephant', 'bear', 'giraffe' & 216 &  117 & 111 & 74 \\ 
  \hline

\end{tabular}
\end{table*}

\begin{table}[t]
\centering
\caption{Class distribution of object-property-based descriptions in the datasets.}
\label{tab:object_description}
\begin{tabular}{c|c|c|c} 
    \hline
    & & \textbf{MIT1003} & \textbf{EMOd} \\ \cline{3-4}     
    \textbf{Category} & \textbf{Label} & \textbf{Object Count} & \textbf{Object Count} \\
    \hline
    Object-color-
    & Red & \bf 2,290 & 576 \\
    hue-based
    & Green & 440 & 210 \\ 
    & Blue & 1,036 &  \bf 1,497 \\ 
    & Gray Scale & 48 & 29 \\ 
    \hline
    Object-color-
    & Low & 910 & 513 \\
    intensity-based
    & Medium & \bf 2,807 & \bf 1,753\\
    & High & 97 & 46 \\ 
    \hline
    Object-scale-
    & Small & 1,259 & 765 \\
    based
    & Medium & 1,258 & 763 \\
    & Large & \bf 1,297 & \textbf{784} \\ 
    \hline
\end{tabular}
\end{table}

\paragraph{\bf MIT1003 Dataset}
The MIT1003 dataset~\cite{judd2009learning} consists of the eye movement data of 15 users on 1,003 static 2D stimuli, including landscapes, indoor scenes, photographs, synthetic images and urban environments, presented in various resolutions. The gaze data was recorded using an ETL 400 ISCAN (240Hz) eye-tracking device. The dataset includes fixation maps, which represent the spatial distribution of gaze fixations for each image. Each participant's scanpath is recorded as a series of fixations of $(x, y)$ points of the screen, with their respective start and end timestamps. For this study, we utilize the MIT1003 version released within the FixaTons repository~\cite{zanca2018fixatons}, which offers a standardized directory structure usable for scripting. The repository includes a supporting Jupyter notebook and library, enabling reliable and reproducible data access.

For color-based image categories, we observed that the intensity minimum ($I_{min}$), maximum ($I_{max}$), and average ($I_{mid}$) values for the MIT1003 dataset were 7, 241, and 124, respectively. Hence, the color-intensity-based categories are determined based on the peak intensity $I_{peak}$ of the images:
\begin{enumerate}
    \item Low-intensity: $I_{peak}<66$
    \item Medium-intensity: $66\le I_{peak} \le 180$
    \item High-intensity: $I_{peak}>180$
\end{enumerate}
The distribution of color-intensity-based and color-hue-based categories, given in Table~\ref{tab:color_classes}, shows that the dominant classes were of medium intensity and red-channel, respectively.

For object-based image categories derived from the outcomes of YOLOv5 model on the datasets, 269 images were detected with no objects. These images were removed from visual analysis, which depends on information of the detected bounding boxes. In the remaining images, a total of 3,814 objects with 74 unique labels have been detected in the dataset, which are described in Table~\ref{tab:object_distribution}, along with their property distribution in Table~\ref{tab:object_description}.

\paragraph{\bf EMOtional attention dataset (EMOd)}
The EMOd dataset~\cite{fan2018emotional} consists of 1,019 emotion-eliciting images accompanied by eye-tracking data of 16 users and detailed annotations at both the object and image levels. Object-level annotations included the contour of the object, along with object name. Image-level annoations had a label indicating the sentiment evoked by the images. For the purpose of our study, we derive our own image-level and object-level annotations, as explained in Section~\ref{sec:implementation}. The images in EMOd are sourced from ``The International Affective Picture System'' (IAPS), which is a set of 321 motion-evoking photos, designed to elicit emotional responses. However, due to restricted public accessibility of the dataset, the IAPS images were excluded from our analysis and so, our analysis is conducted on the remaining 698 publicly available images. 

For color-intensity-based image categories, we observed that $I_{min}$, $I_{max}$, and $I_{mid}$ values give:
\begin{enumerate}
    \item Low-intensity: $I_{peak}<62$
    \item Medium-intensity: $62\le I_{peak} \le 165$
    \item High-intensity: $I_{peak}>165$
\end{enumerate}
The distribution of color-intensity- and color-hue-based categories, given in Table~\ref{tab:color_classes}, shows that the dominant classes were of medium intensity and blue-channel, respectively.

YOLOv5 model detected 2,312 unique objects in the dataset, whose categories are given in Table~\ref{tab:object_distribution} along with property descriptions in Table~\ref{tab:object_description}. All processing steps are followed homogeneously for both datasets.

\subsection{Implementation} \label{sec:implementation}
EnsembleGaze is implemented on AMD Ryzen 5 9600X 6-Core Processor, with a 16GB RAM with a 8GB Nvidia GeForce RTX 5060 GPU. All coding for analysis and visualizations were implemented in Python 3.14, using libraries, like scipy, scikit-learn, numpy, pandas, sklearn, mathplotlib and Plotly, using PyCharm IDE. 

It must be noted that the MIT1003 dataset has finer and richer data than the EMOd. This leads to three specific changes in running EnsembleGaze for the case study of EMOd. Firstly, owing to the absence of scanpath data, the user-clustering based on scanpath-distance-based features cannot be implemented. Secondly, the original EMOd dataset is released as MATLAB\textsuperscript{\textregistered} files, where the each scanpath of a (user, stimulus) data point is given in the format $(x, y, FD)$. Here, we assume that the $TTF$ is considered at the timestamp 0. Since there is no significant variability of $TTF$ between data points, the $TTF$ value is ignored for all computations in EMOd dataset. Thirdly, for each (user, stimulus) scanpath and the fixation durations ($\mathbf{FD}$) provided, we compute the start and end timestamps of each fixation, with an assumption that the $TTF$=0. This helps in organizing the scanpaths as an array of fixations, with each tuple having the schema $(x, y, start\_time, end\_time)$, that is followed in MIT1003. This standardizes the input data format to EnsembleGaze.

\subsection{Experiments} \label{sec:expt}
\begin{table*}[htp]
\centering
\caption{Summary of experiments along with the setup and features for the case studies}
\label{tab:experiments}
\scriptsize
\begin{tabular*}{\textwidth}{@{\extracolsep{\fill}}c|c|c|c|p{0.5\textwidth}}
\hline
\textbf{ID} & \textbf{Experiment Outcome}
& \textbf{MIT1003} & \textbf{EMOd} 
& \textbf{Setup Specifications} \\ 
\hline

\expt{1}
& User-clustering
& \checkmark & $\times$
&
\begin{itemize}
\item Feature type: Scanpath-distance-based features
\item Experiment run: on each distance metrics --
  \begin{enumerate}
  \item Euclidean
  \item Fr\'echet
  \item Hausdorff
  \item Dynamic Time Warping (DTW)
  \item String-based Time Delay Embedding (SBTDED)
  \end{enumerate}
\item Feature matrix: $\mathbf X_d\in\mathbb{R}^{(N\times N)}$
  \begin{itemize}
    \item $N$=15 for MIT1003
  \end{itemize}
\end{itemize}

\\
\hline

\expt{2}
& User- \& Stimuli-clustering
& \checkmark & \checkmark
&
\begin{itemize}
\item Feature type: Fixation-based features, based on 5 statistical measures of distributions of \ie~$FC$, $TTF$, $\mathbf{FD}$
  \begin{itemize}
  \item $TTF$ data is not available for EMOd, hence the feature vector and feature matrix are truncated
  \end{itemize}
\item Experiment run:
  \begin{enumerate}
  \item User-clustering
  \item Stimuli-clustering
  \end{enumerate}
\item Feature matrix: $\mathbf X_D\in (\mathbb{R}^+)^{(\mathcal{N}\times 5N_F)}$
  \begin{itemize}
  \item For each run:
    \begin{enumerate}
    \item $\mathcal{N}$=$N$=15, $N_F$=3 for MIT1003; $\mathcal{N}$=$N$=16, $N_F$=2 for EMOd.
    \item $\mathcal{N}$=$S$=1,003, $N_F$=3 for MIT1003; $\mathcal{N}$=$S$=698, $N_F$=2 for EMOd.
    \end{enumerate}
  \end{itemize}
\end{itemize}
\\
\hline

\expt{3}
& User-clustering
& \checkmark & \checkmark
&
\begin{itemize}
\item Feature type: An extension to the user-clustering in \expt{2}, with the following differences:
  \begin{itemize}
  \item Feature vector is computed for each object-property-based category 
  \item All the category-based feature vectors are concatenated to give the final feature vector
  \end{itemize}
\item Experiment run: on each image-property --
  \begin{enumerate}
  \item Color-hue: Red/Blue/Green/Greyscale, \ie~4 categories
  \item Color-intensity: Low/Medium/High, \ie~3 categories
  \item Object-Scale: Small/Medium/Large, \ie~3 categories
  \item Hue+Intensity+Scale, \ie~10 categories
  \end{enumerate}
  
\item Feature matrix: $\mathbf X_D\in (\mathbb{R}^+)^{(N\times 5N_F)}$
  \begin{itemize}
  \item $N$=15 for MIT1003; $N$=16 for EMOd
  \item For each run:
  \begin{enumerate}
  \item $N_F$=3$\times$4=12 for MIT1003; $N_F$=2$\times$4=8 for EMOd
  \item $N_F$=3$\times$3=9 for MIT1003; $N_F$=2$\times$3=6 for EMOd
  \item $N_F$=3$\times$3=9 for MIT1003; $N_F$=2$\times$3=6 for EMOd
  \item $N_F$=3$\times$10=30 for MIT1003; $N_F$=2$\times$10=20 for EMOd  
  \end{enumerate}
  \end{itemize}
\end{itemize}
\\
\hline

\expt{4}
& (User, stimuli)-clustering
& \checkmark & \checkmark
&
\begin{itemize}
\item Feature type: Same as \expt{2}
\item Experiment run:
  \begin{enumerate}
  \item Consensus subspace clustering with user-guided sequence
  \item Consensus subspace clustering with stimulus-guided sequence
  \item Biclustering with user-guided unfolding
  \item Biclustering with stimulus-guided unfolding
  \end{enumerate}
\item Feature tensor: $\mathcal X\in (\mathbb{R}^+)^{(N\times S\times 5)}$
  \begin{itemize}
  \item $N$=15, $S$=1,003 for MIT1003; $N$=16, $S$=698 for EMOd
  \end{itemize}
\end{itemize}
\\
\hline
\end{tabular*}
\end{table*}

\begin{table*}[htp]
\centering
\caption{Optimal $k$ value for baseline clustering input across all experiments, and the metric values, namely, Silhouette index (SI), Davies Bouldin index (DB), and Calinski Harabasz index (CH), used to finalize the $k$ value. Colors \textcolor{red}{red} and \textcolor{blue}{blue} highlight the worst and the best values, respectively, to indicate the range for each metric for a specific dataset.}
\label{tab:optimal-k}
\scriptsize
\begin{tabular}{c|c|c||c|c|c|c||c|c|c|c}
\hline

&&
& \multicolumn{4}{c||}{\textbf{MIT1003}} 
& \multicolumn{4}{c}{\textbf{EMOd}} \\\cline{4-11}

\bf ID & \bf Outputs & \bf Expt. Setup
& \textbf{SI}$\uparrow$ & \textbf{DB}$\downarrow$
& \textbf{CH}$\uparrow$ & $\mathbf{k}$
& \textbf{SI}$\uparrow$ & \textbf{DB}$\downarrow$
& \textbf{CH}$\uparrow$ & $\mathbf{k}$ \\\hline

\expt{1} & User-clustering &
&&&& &&&& \\
& \multicolumn{1}{r|}{Distance-metric:} & Euclidean Distance
& \textcolor{red}{\textbf{0.036}} & 0.88 & \textcolor{red}{\bf 2.12} & 9
&  &  &  &  \\
&& Fr\'echet Distance
& 0.056 & 1.2 & 2.29 & 7
&  &  &  &  \\
&& Hausdorff Distance
& 0.12 & 1.18 & 3.49 & 3
& - & - & - & - \\
&& DTW
& 0.05 & 1.02 & 2.18 & 8
&  &  &  &  \\
&& SBTDED
& 0.04 & 1.28 & 2.37 & 6
&  &  &  &  \\
\hline

\expt{2} & User-clustering & -
& \textcolor{blue}{\textbf{0.62}} & 0.35 & 34.1 & 3
& \textcolor{blue}{\textbf{0.65}} & \textcolor{blue}{\textbf{0.38}} & 86.27 & 3 \\

& Stimuli-clustering & -
& 0.47 & 0.61 & \textcolor{blue}{\textbf{1,598.12}} & 4
& 0.56 & 0.57 & \textcolor{blue}{\textbf{1,795.37}} & 4 \\
\hline

\expt{3} & User-clustering &
&&&& &&&& \\
& \multicolumn{1}{r|}{Image-property:} & Color-hue Category Features
& 0.56 & 0.36 & 49.12 & 4
& 0.39 & 0.86 & 20.47 & 3 \\

&& Color-intensity Category Features
& 0.57 & \textcolor{blue}{\textbf{0.34}} & 46.02 & 4
& \textcolor{red}{\textbf{0.23}} & 0.74 & 9.82 & 6\\

&& Scale Category Features
& 0.56 & \textcolor{blue}{\textbf{0.34}} & 46.83 & 4
& 0.26 & 0.941 & 8.40 & 4 \\

&& Hue+Intensity+Scale Category Features
& 0.56 & 0.35 & 47.07 & 4 & 0.28 & 0.84 & 8.37 & 5 \\
\hline

\expt{4} & (User, stimuli)-clust. &
&&&& &&&& \\
& \multicolumn{1}{r|}{CSC:} & User-guided Sequence
& 0.5 & 0.66 & 11.51 & (2, 2)
& 0.46 & 0.86 & 11.41 & (2, 2) \\
&& Stimulus-guided Sequence
& 0.37 & 1.06 & \textcolor{blue}{\textbf{537.38}} & (2, 2)
& 0.33 & 1.32 & 300.93 & (2, 2) \\

& \multicolumn{1}{r|}{Biclustering:} & User-guided Unfolding
& 0.45 & 0.88 & 11.12 & (2, 2)
& 0.25 & 1.25 & \textcolor{red}{\bf 6.98} & (2, 2) \\
&& Stimulus-guided Unfolding
& 0.30 & \textcolor{red}{\textbf{1.21}} & \textcolor{blue}{\textbf{422.07}} & (2, 2)
& \textcolor{red}{\textbf{0.23}} & \textcolor{red}{\textbf{1.51}} & 220.89 & (2, 2) \\ 
\hline
\end{tabular}

\vspace{0.2em}
\textit{Key:-} CSC: Consensus Subspace Clustering
\end{table*}

The objectives of the experiments are:
\begin{description}
\item[\expt{1}:]determines the optimal distance metric for consensus clustering of users.
\item[\expt{2}:]examines the results of consensus clustering based on fixation-based features for both user- and stimuli-clustering.
\item[\expt{3}:]examines the influence of different image-property-based categories on the user-clustering, where the feature vectors are computed for each category and a concatenated feature vector is used for clustering. This is an extension to the user-clustering in \expt{2} using longer (concatenated) feature vectors.
\item[\expt{4}:]compares the two different high-dimensional clustering techniques.
\end{description}

A summary of experiments used in each case study, along with the required setup, is given in Table~\ref{tab:experiments}. The features required for the experiments are computed as described in Section~\ref{sec:feature_selection}. 

Three baseline clustering methods are chosen for consensus clustering, namely $k$-means++, hierarchical agglomerative and spectral clustering methods. The optimal number of clusters $k$, needed as input for any clustering method within each experiment, is decided using three metrics, namely, Silhouette index, Davies Bouldin index, and Calinski Harabasz index. For both user- and stimuli-clustering experiments, a range of $k$ values were tested. We choose the range of \{2,10\}, where beyond the upper limit of 10 implies highly fragmented clusters, especially with the size of data points in the chosen datasets. We take the consensus of the three scores to determine the optimal $k$. In the event of no consensus, several runs with different seeds help determine optimal $k$. After systematic analysis, we find optimal $k$=3 for user-clustering, and optimal $k$=4 for stimuli-clustering, for both datasets. Table~\ref{tab:optimal-k} gives the summary of our findings.

\subsection{Case Study 1: MIT1003 Dataset} \label{sec:mit1003}
\begin{table*}[tp]
    \centering
    \caption{Cluster quality and comparison metrics for \expt{1}, \expt{2}, \expt{3} and \expt{4} for MIT1003.}
    \label{tab:metrics-mit1003}
    \scriptsize
    \begin{tabular}{ m{2cm} | m{3cm} | m{1.4cm}| m{1.2cm} | m{1.5cm} | m{1.2cm} | m{1cm} | m{1cm}}
  \hline
  \textbf{Summary} 
  & \textbf{Comparison}
  & \textbf{Cohen's $\kappa$}$\uparrow$
  & \textbf{Accuracy}$\uparrow$
  & \textbf{MCC}$\uparrow$
  & \textbf{F1 Score}$\uparrow$ 
  & \textbf{ARI}$\uparrow$ 
  & \textbf{NMI}$\uparrow$ \\
  \hline
  \multirow{4}{3cm}{User-clustering} 
  & Hierarchical vs. Consensus & 0.42 & 0.6 & 0.55 & 0.63 & 0.34 & 0.61 \\
  & Spectral vs. Consensus & \textcolor{blue}{\textbf{1.00}} & \textcolor{blue}{\textbf{1.00}} & \textcolor{blue}{\textbf{1.00}} & \textcolor{blue}{\textbf{1.00}} & \textcolor{blue}{\textbf{1.00}} & \textcolor{blue}{\textbf{1.00}} \\
  & KMeans++ vs. Consensus & 0.43 & 0.67 & 0.49 & 0.62 & 0.35 & 0.52 \\
  & \expt{3}'s all features vs. Consensus & 0.55 & 0.73 & 0.64 & 0.7 & 0.41 & 0.64 \\
  \hline
  \hline
  \multirow{6}{3cm}{Stimuli-clustering} 
  & Community vs. Consensus & 0.96 & 0.97 & 0.96 & 0.96 & 0.92 & 0.9 \\
  & Hierarchical vs. Consensus & 0.68 & 0.77 & 0.7 & \textcolor{blue}{\textbf{0.77}} & 0.54 & 0.62 \\
  & Spectral vs. Consensus & \textcolor{blue}{\textbf{0.76}} & \textcolor{blue}{\textbf{0.85}} & \textcolor{blue}{\textbf{0.77}} & 0.7 & \textcolor{blue}{\textbf{0.8}} & \textcolor{blue}{\textbf{0.84}} \\
  & KMeans++ vs. Consensus & 0.71 & 0.81 & 0.72 & 0.65 & 0.65 & 0.68\\
  & Colour vs. Consensus & 0.06 & \textcolor{red}{\textbf{0.36}} & 0.07 & 0.25 & \textcolor{red}{\textbf{0.008}} & \textcolor{red}{\textbf{0.008}} \\
  & Intensity vs. Consensus & \textcolor{red}{\textbf{0.03}} & 0.41 & \textcolor{red}{\textbf{0.04}} & \textcolor{red}{\textbf{0.18}} & 0.01 & 0.006\\
  \hline
  \hline
  \multirow{2}{3cm}{High-dimensional Clustering} 
  & User-guided CSC vs. User-guided Biclustering & \textcolor{blue}{\textbf{0.76}} & \textcolor{blue}{\textbf{0.93}} & \textcolor{blue}{\textbf{0.79}} & \textcolor{blue}{\textbf{0.88}} & \textcolor{blue}{\textbf{0.68}} & \textcolor{blue}{\textbf{0.59}} \\
  & Stimuli-guided CSC vs. Stimuli-guided Biclustering & 0.73 & 0.88 & 0.73 & \textcolor{blue}{\textbf{0.87}} & 0.57 & 0.44 \\
  \hline
    \end{tabular}

\vspace{0.2em}
\textit{Key:-} CSC: Consensus Subspace Clustering
\end{table*}

\begin{figure*}
    \centering
    \includegraphics[width=\linewidth]{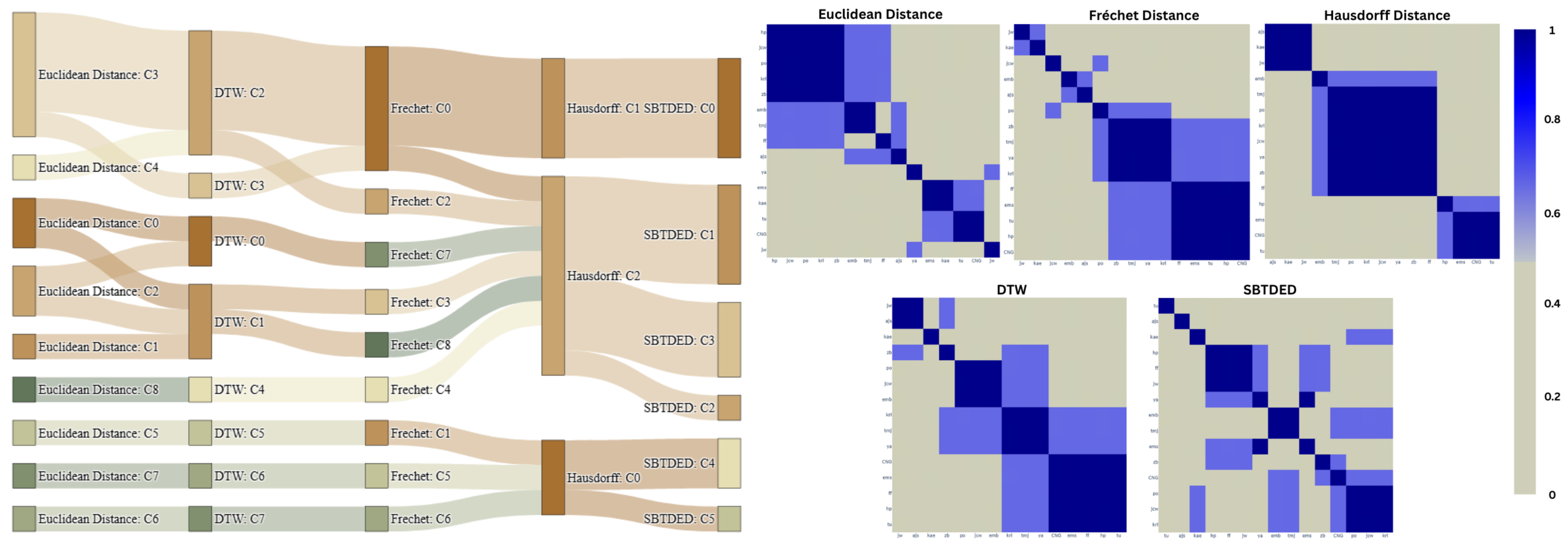}
    \caption{Results of experiment \expt{1} for MIT1003 dataset. The plot on the left shows an alluvial chart of clusterings using the 5 different distance metrics. The plot on the right shows the coassociation matrices of the ensemble cluster for each distance metric.}
    \label{fig:e1_mit}
\end{figure*}

\begin{figure*}
    \centering
    \includegraphics[width=\linewidth]{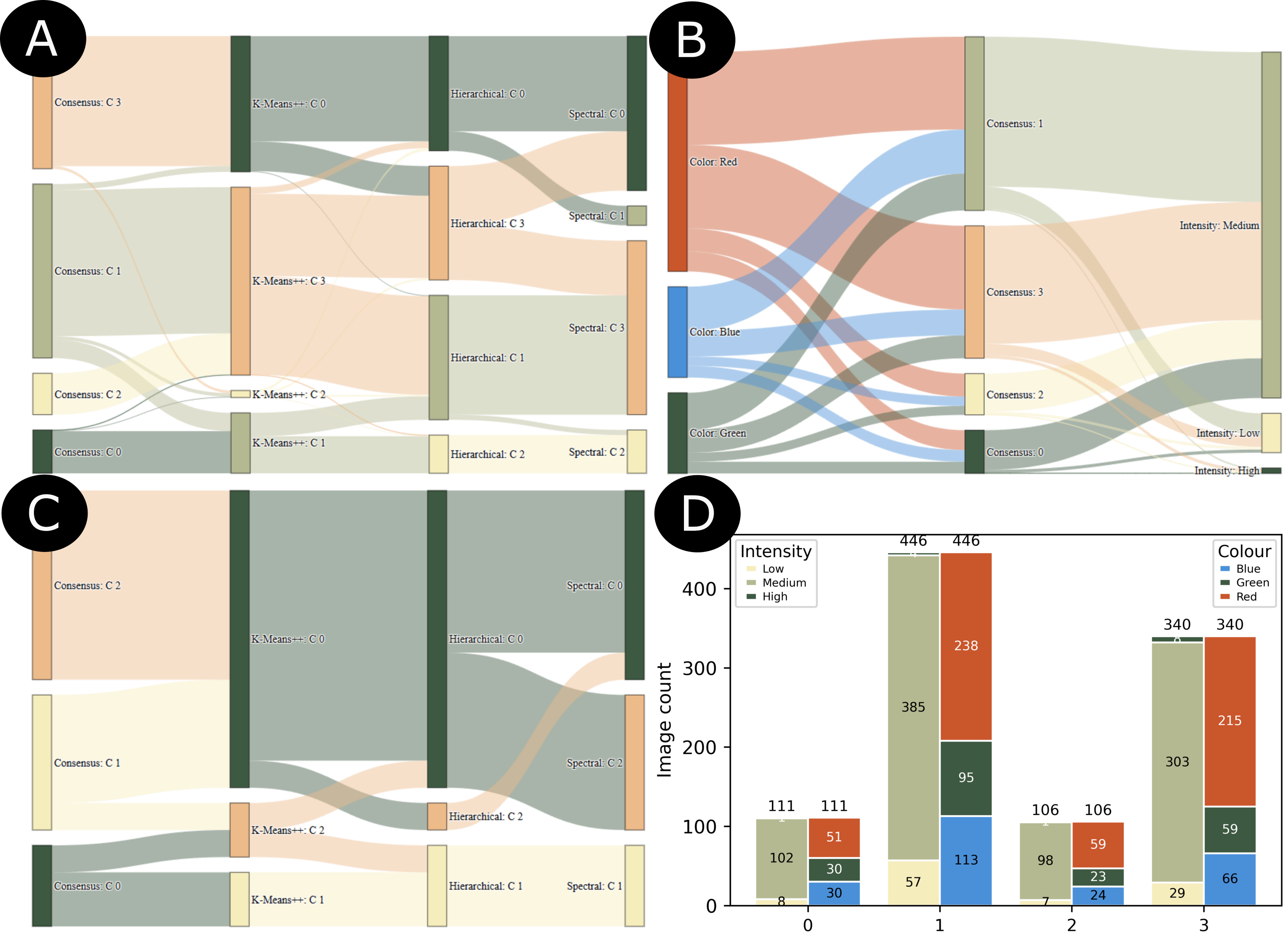}
    \caption{Results of experiment \expt{2} for MIT1003 dataset. Plot -- and -- show the results of stimuli-clustering, and comparing stimuli-clustering with image categories in alluvial charts. The composition of stimuli-clustering partitions in terms of image properties of color hue and intensity are shown in plot --. User-clustering is shown in plot --, with the results of baseline clustering schemes. }
    \label{fig:e2_mit}
\end{figure*}

\begin{figure*}
    \centering
    \includegraphics[width=\linewidth]{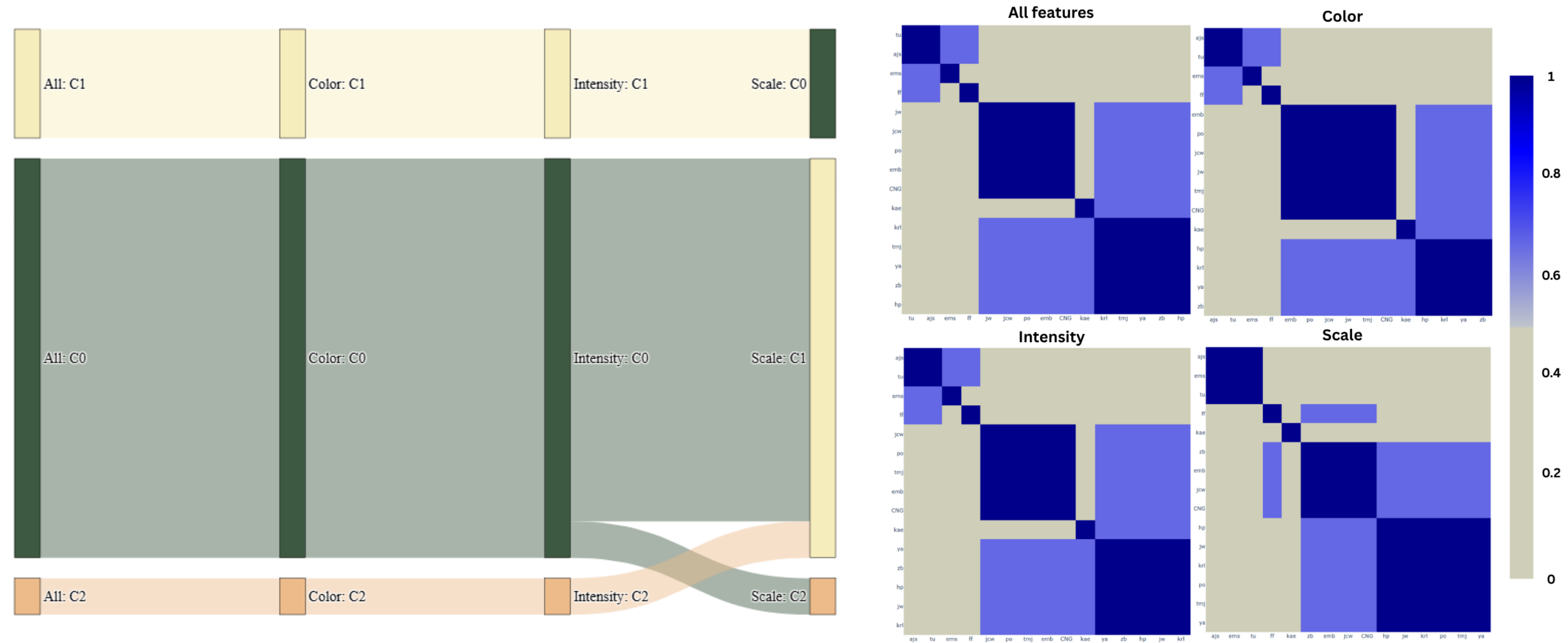}
    \caption{Results of experiment \expt{3} for MIT1003 dataset. The alluvial plot (left) shows the user clustering results for different types of object-property-based feature vectors were passed to EnsembleGaze. The panel of co-association matrices (right) visualizes the ensemble clustering outcome for each feature category.}
    \label{fig:e3_mit}
\end{figure*}

\begin{figure*}
    \centering
    \includegraphics[width=0.85\linewidth]{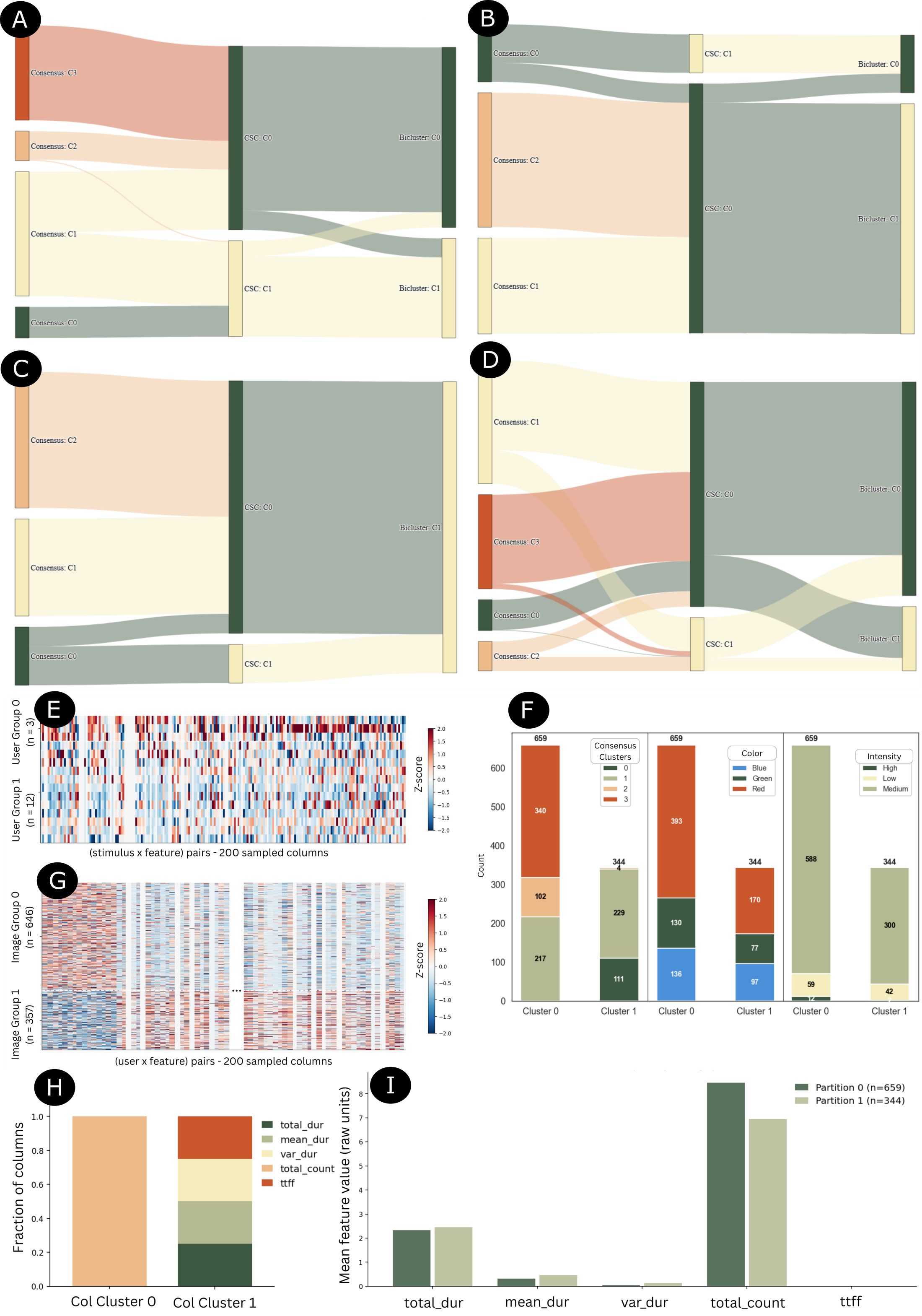}
    \caption{Experiment \expt{4} results for the MIT1003 dataset. A and C show first- and second-level of stimulus-guided CSC and biclustering alluvial plots respectively, compared with \expt{2} results. B and D show first- and second-level user-guided configurations respectively, compared with \expt{2} consensus results. E and G show reordered activation heatmaps for user- and stimulus-guided biclustering. F show cluster composition of the two stimulus-guided sequence CSC clusters using \expt{2} data and image categories. H and I show feature proportions per cluster partition for stimulus-guided biclustering and CSC, respectively.}
    \label{fig:e4_mit}
\end{figure*}

\paragraph{Experiment \expt{1}} 
The optimal number of clusters for each distance metric is initialized according to Table~\ref{tab:optimal-k}. Figure~\ref{fig:e1_mit}A shows the cluster members for each scheme. Visual analysis of the seriated coassociation matrices provides evidence of 3–5 clusters among the 15 users. The Hausdorff distance matrix yielded a lower number of clusters compared to the other methods. Notably, the Hausdorff distance also achieved the best silhouette score and CH index among the five distance metrics tested. In contrast, the partitions produced by Euclidean and Fr\'echet distances show several single-member clusters. The Euclidean distance metric also received the lowest silhouette score and CH index across experiments compared to all other schemes. Meanwhile, the coassociation matrices for DTW and SBTDED do not exhibit clear partitions. Overall, the distance-metric-based schemes received poorer silhouette scores, CH indices, and DB indices compared to Experiments \expt{2}, \expt{3}, and \expt{4}, whereas the Hausdorff distance showed clear clustering patterns upon visual inspection.

\paragraph{Experiment \expt{2}} 
For the stimuli-clustering of Experiment \expt{2}, the obtained consensus labels are compared with baseline clustering methods in Figure~\ref{fig:e2_mit}A. Upon visual inspection, cluster membership varies across each scheme, with the spectral method yielding the highest correlation with the consensus voting method. This trend is also reflected in Table~\ref{tab:metrics-mit1003}, where the spectral and consensus methods exhibit high Cohen's kappa and MCC values. The composition of image and object properties for each cluster is shown in Figure~\ref{fig:e2_mit}B and Figure~\ref{fig:e2_mit}D.

The user-clustering of Experiment \expt{2} achieved the best silhouette score and DB index across all experiments. The three consensus clustering partitions aligned highly with the spectral clustering results, as evidenced by the perfect Cohen's \(\kappa \), accuracy, MCC, and F1 score in Table~\ref{tab:metrics-mit1003}.

\paragraph{Experiment \expt{3}} 
The user clustering obtained by utilizing all image-property-based feature vectors is shown in Figure~\ref{fig:e3_mit}. The alluvial plot demonstrates that the user partitions determined by color and intensity align exactly with the partitions discovered when using all features, whereas two users' memberships shift when scale-based features are used. We observe 3–4 clusters upon visual inspection of the coassociation matrices for all categories.

The alignment between the consensus clustering of Experiments \expt{2} and \expt{3} is not identical, as indicated by the Cohen's \(\kappa \) of 0.55 in Table~\ref{tab:metrics-mit1003}. However, this comparison still maintains a high accuracy and F1 score, which is discussed further.

\paragraph{Experiment \expt{4}} 
Stimulus-guided unfolding and stimulus-guided sequential CSC achieved high CH index values for two stimuli partitions and two user partitions in Table~\ref{tab:optimal-k}, indicating clear separability between the partitions. Figure~\ref{fig:e4_mit}A shows the comparative clustering results of Experiment \expt{2}'s stimuli clustering alongside the stimulus-guided unfolding biclusters and stimulus-guided sequential CSC. Except for a small population, the two stimulus-guided high-dimensional clustering schemes exhibit general alignment, which is visible in the alluvial diagram. The accuracy and F1 score comparing the two stimulus-guided schemes also indicate high correspondence. Because the stimulus-guided sequential CSC demonstrated a better CH index, we visualize the composition of the two stimuli partitions with respect to Experiment \expt{2}'s consensus stimuli clustering and image categories in Figure~\ref{fig:e4_mit}F. Additionally, upon visually examining the dominant feature proportions of each stimulus-guided partition (Figures~\ref{fig:e4_mit}H and ~\ref{fig:e4_mit}I), plot H reveals that bicluster C0 is composed of members with a high \(\mu(FC)\), whereas C1 is driven by fixation duration and TTF features. Furthermore, the CSC partition C0 is the larger partition and is predominantly represented by a higher total fixation count, corresponding with bicluster C0.

The stimulus-guided biclustering also exhibits a clear checkerboard pattern in the reordered activation maps (Figure~\ref{fig:e4_mit}G), which indicates high separability between the stimuli partitions. From the stimulus-guided high-dimensional clustering schemes, user partitions are derived using majority voting, and the results are visualized in Figure~\ref{fig:e4_mit}C, comparing the outcome with the user clusters of Experiment \expt{2}. The biclustering produces a single user partition due to the tie-breaking condition of the majority voting, where a user is mapped to cluster 1 if they belong to both bicluster partitions. This indicates that a significant number of users belong to both biclusters. On the other hand, the user partitions of the stimulus-guided CSC show two clusters: one with thirteen members and the other with two members. The thirteen-member cluster completely encompasses the results of two clusters, C1 and C2, from Experiment \expt{2}'s consensus user clustering.

In contrast, user-guided biclustering does not show a clear checkerboard pattern in the reordered activation maps in Figure~\ref{fig:e4_mit}E. This indicates that the clustering is not as clearly separable, which is further evidenced by the comparatively lower CH index. Notably, the CH index of the user-guided CSC is also on the lower side, closely resembling the user-guided biclustering results. Comparing the outcomes of the user-guided high-dimensional clustering schemes in Figure~\ref{fig:e4_mit}B, the CSC identifies two partitions of thirteen and two members, while the biclustering shows a twelve-three divide. Comparing these to the results of the Experiment \expt{2} consensus user clustering, it is apparent that two of the clusters (C1 and C2) and one user from C0 constitute the thirteen users in the CSC. This heavily resembles plot C, which maps the user partitions of the stimulus-guided high-dimensional clustering. Furthermore, the stimuli partitions of the user-guided schemes are compared in the alluvial chart (Figure~\ref{fig:e4_mit}D). While two clusters are present in both the CSC and biclustering, a large portion of CSC's C1 merges with bicluster C0.

\subsection{Case Study 2: EMOd Dataset} \label{sec:emod}
\begin{table*}[t]
    \centering
    \caption{Cluster quality and comparison metrics for \expt{2}, \expt{3} and \expt{4} for EMOd.}
    \label{tab:metrics-emod}
    \scriptsize
    \begin{tabular}{ m{2cm} | m{3cm} | m{1.4cm}| m{1.2cm} | m{1.5cm} | m{1.2cm} | m{1cm} | m{1cm}}
  \hline
  \textbf{Summary} 
  & \textbf{Comparison}
  & \textbf{Cohen's $\kappa$}$\uparrow$
  & \textbf{Accuracy}$\uparrow$
  & \textbf{MCC}$\uparrow$
  & \textbf{F1 Score}$\uparrow$ 
  & \textbf{ARI}$\uparrow$ 
  & \textbf{NMI}$\uparrow$ \\
  \hline
  \multirow{3}{3cm}{User-clustering} 
  & Hierarchical vs. Consensus & 0.48 & 0.69 & 0.5 & 0.64 & 0.18 & 0.31 \\
  & Spectral vs. Consensus & 0.59 & 0.75 & \textcolor{blue}{\textbf{0.65}} & 0.7 & 0.31 & \textcolor{blue}{\textbf{0.48}} \\
  & KMeans++ vs. Consensus & \textcolor{blue}{\textbf{0.62}} & \textcolor{blue}{\textbf{0.81}} & 0.62 & \textcolor{blue}{\textbf{0.84}} & \textcolor{blue}{\textbf{0.4}} & 0.45 \\
  & \expt{3}'s all features vs. Consensus & 0.35 & 0.56 & 0.4 & 0.53 & 0.10 & 0.29 \\
  \hline
  \hline
  \multirow{5}{3cm}{Stimuli-clustering}
  & Hierarchical vs. Consensus & 0.40 & 0.59 & 0.42 & 0.52 & 0.34 & 0.42 \\
  & Spectral vs. Consensus & 0.37 & 0.54 & 0.38 & 0.6 & 0.25 & 0.4 \\
  & KMeans++ vs. Consensus & \textcolor{blue}{\textbf{0.75}} & \textcolor{blue}{\textbf{0.82}} & \textcolor{blue}{\textbf{0.76}} & \textcolor{blue}{\textbf{0.82}} & \textcolor{blue}{\textbf{0.57}} & \textcolor{blue}{\textbf{0.58}}\\
  & Colour vs. Consensus & \textcolor{red}{\textbf{0.03}} & \textcolor{red}{\textbf{0.33}} & \textcolor{red}{\textbf{0.03}} & 0.23 & 0.006 & 0.02 \\
  & Intensity vs. Consensus & 0.05 & 0.35 & 0.08 & \textcolor{red}{\textbf{0.22}} & \textcolor{red}{\textbf{0.005}} & \textcolor{red}{\textbf{0.01}} \\
  \hline
  \hline
  \multirow{2}{3cm}{High-dimensional Clustering} 
  & User-guided CSC vs. User-guided Biclustering & \textcolor{red}{\textbf{0.19}} & \textcolor{red}{\textbf{0.63}} & \textcolor{red}{\textbf{0.22}} & \textcolor{red}{\textbf{0.56}} & \textcolor{red}{\textbf{0.02}} & \textcolor{red}{\textbf{0.04}} \\
  & Stimuli-guided CSC vs. Stimuli-guided Biclustering & 0.5 & 0.76 & 0.53 & \textcolor{red}{\textbf{0.56}} & 0.26 & 0.23\\
  \hline
    \end{tabular}

\vspace{0.2em}
\textit{Key:-} CSC: Consensus Subspace Clustering
\end{table*}

\begin{figure*}
    \centering
    \includegraphics[width=\linewidth]{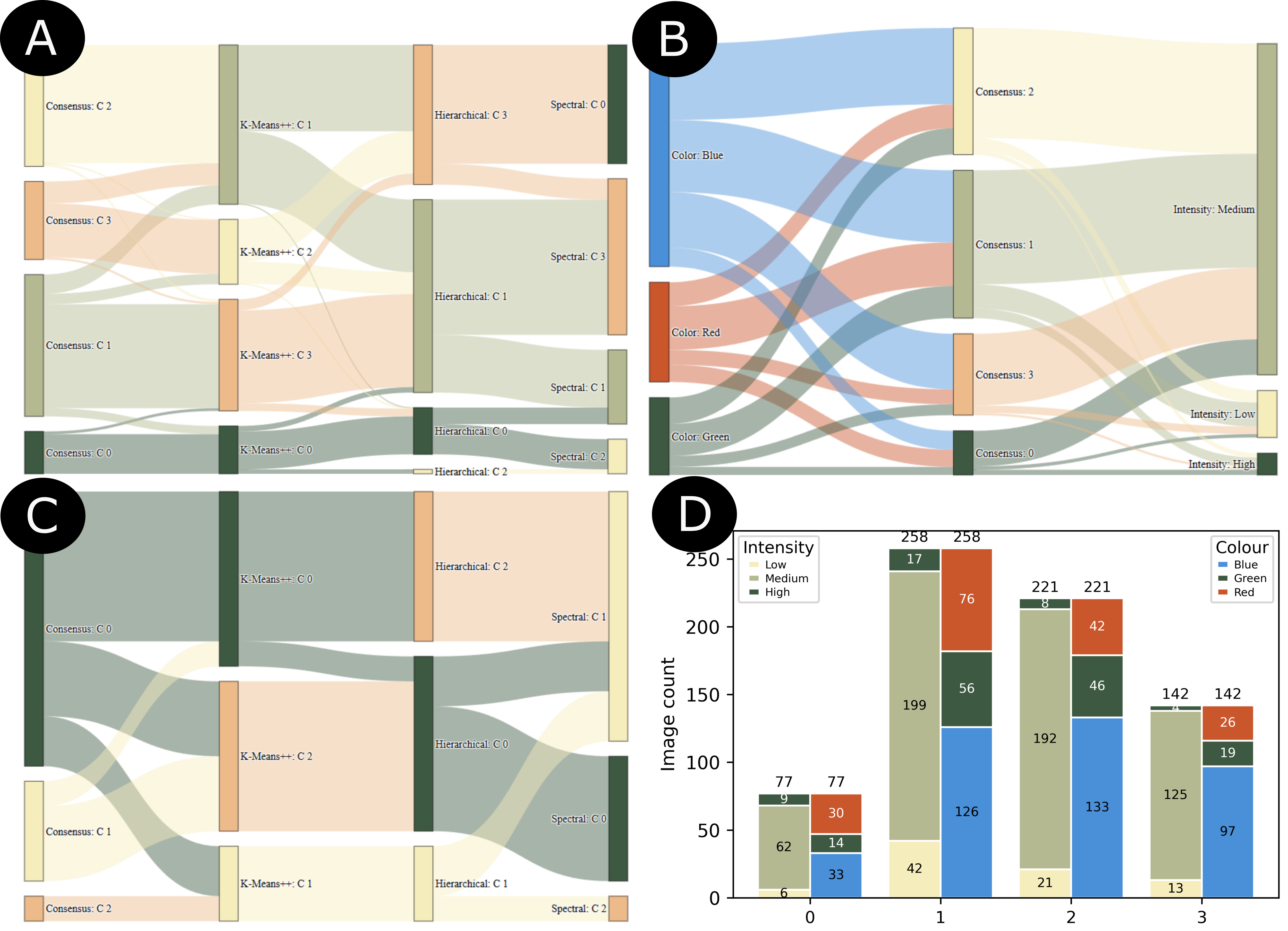}
    \caption{Results of experiment \expt{2} on EMOd dataset. A shows an alluvial chart comparing the outcome of stimuli consensus clustering with that of baseline clustering. B and D shows the composition of stimuli clustering with respect to color hue and intensity of the images. C shows an alluvial chart of the consensus user clustering with that of baseline clustering.}
    \label{fig:e2_emod}
\end{figure*}

\begin{figure*}
    \centering
    \includegraphics[width=\linewidth]{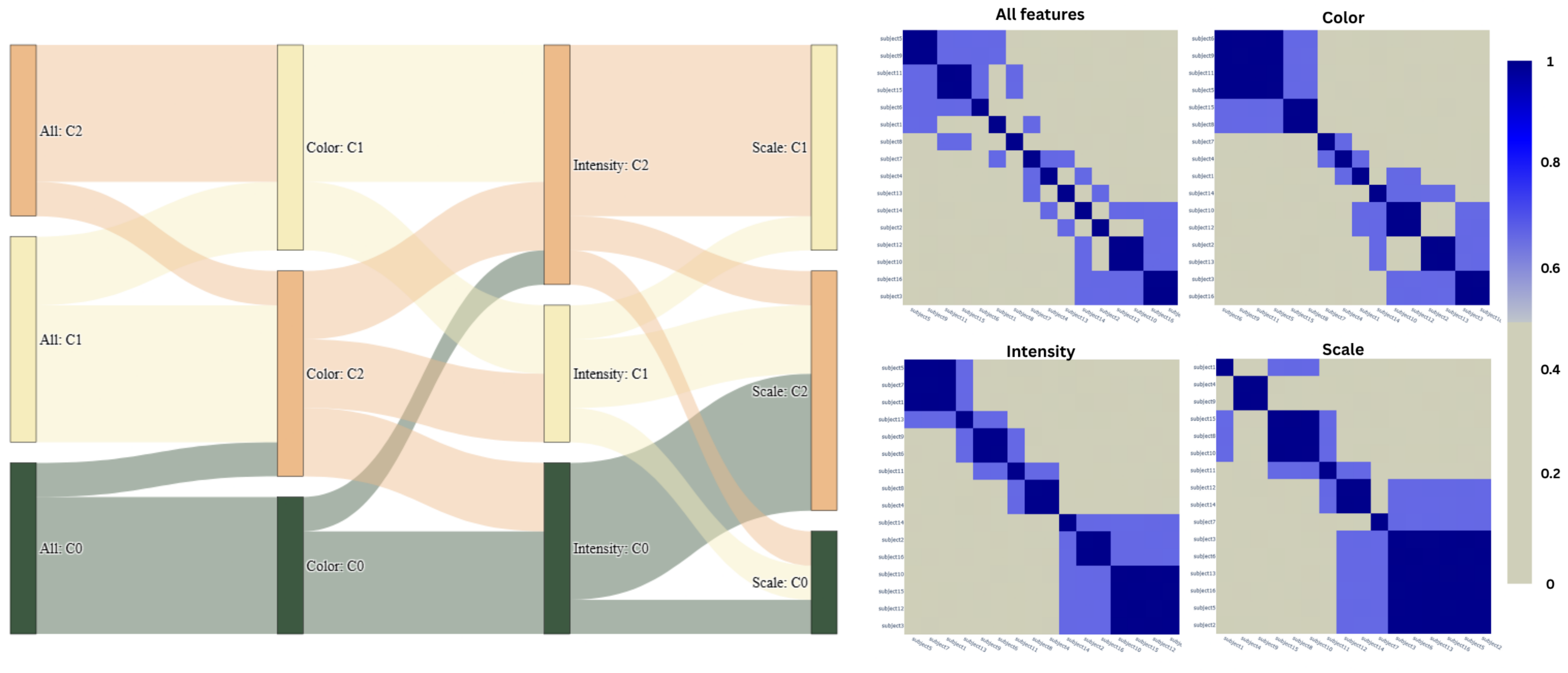}
    \caption{Results of experiment \expt{3} on EMOd dataset. The alluvial chart (left) shows user clustering when object-based fixation features were used for consensus clustering, compared to using a subset of the features based on color hue, object intensity and object scale. The panel of coassociation matrices shows user coassociation for the different types of object-based feature sets. }
    \label{fig:e3_emod}
\end{figure*}

\begin{figure*}
    \centering
    \includegraphics[width=0.85\linewidth]{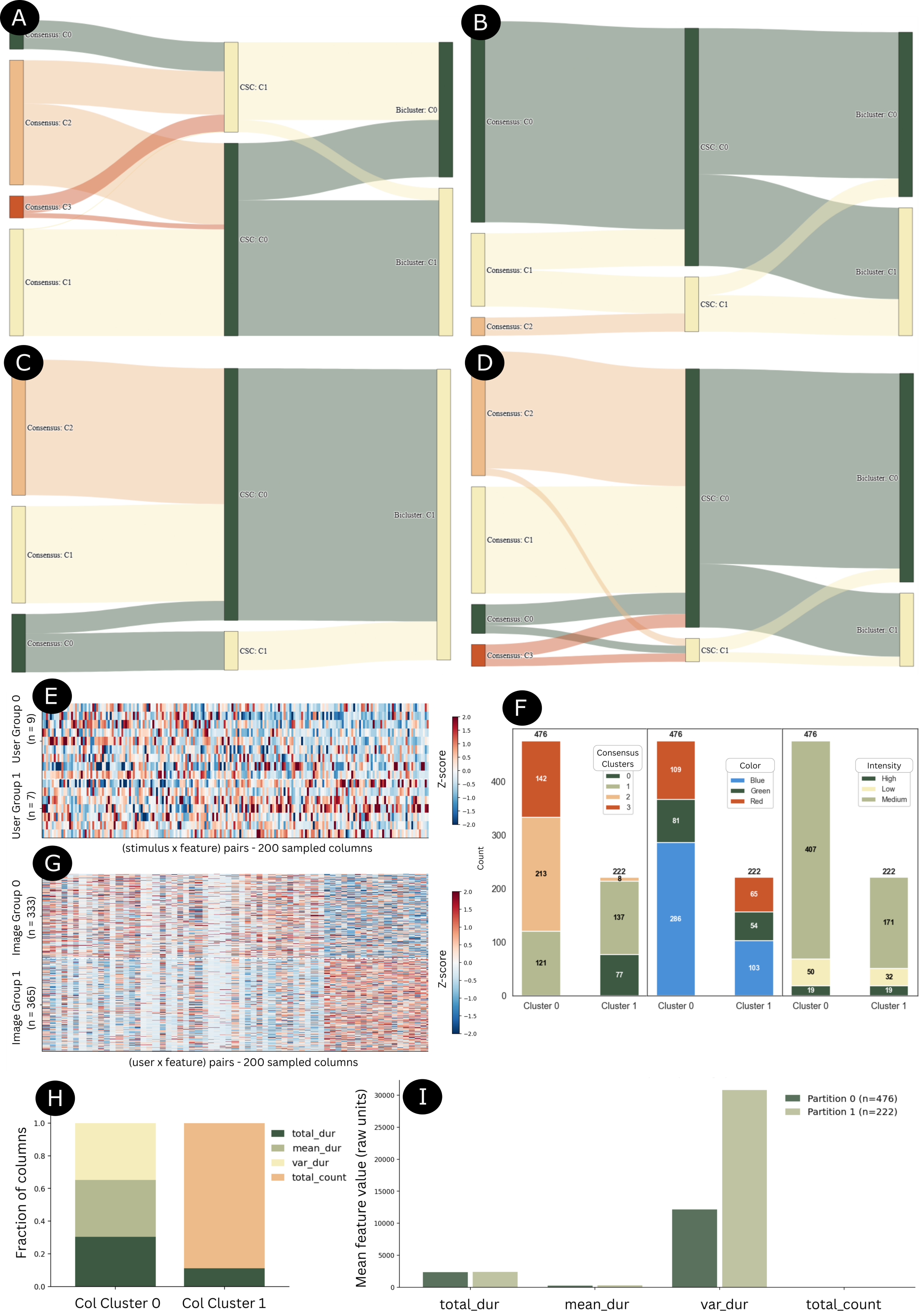}
    \caption{Experiment \expt{4} results for the EMOd dataset. A and C show first- and second-level of stimulus-guided CSC and biclustering alluvial plots respectively, compared with \expt{2} results. B and D show first- and second-level user-guided configurations respectively, compared with \expt{2} consensus results. E and G show reordered activation heatmaps for user- and stimulus-guided biclustering. F show cluster composition of the two stimulus-guided sequence CSC clusters using \expt{2} data and image categories. H and I show feature proportions per cluster partition for stimulus-guided biclustering and CSC, respectively.}
    \label{fig:e4_emod}
\end{figure*}

\paragraph{Experiment \expt{2}}
Table~\ref{tab:optimal-k} demonstrates that both user-clustering and stimuli-clustering achieve robust silhouette scores, DB indices, and CH indices. Furthermore, the alluvial chart (Figure~\ref{fig:e2_emod}A) reveals that consensus clusters C0, C1, and C3 correspond predominantly to unique clusters within KMeans++. This alignment is further supported by the higher metric values yielded in the consensus-versus-KMeans++ comparison in Table~\ref{tab:metrics-emod}. Conversely, image properties such as color and intensity exhibit low correspondence (Figure~\ref{fig:e2_emod}B), as indicated by the lower Cohen's \(\kappa \) and MCC values.

The consensus user-clustering results (Figure~\ref{fig:e2_emod}C) display high correspondence with both the spectral and KMeans++ baseline configurations. This strong alignment is explicitly affirmed by the elevated Cohen's \(\kappa \) and MCC values presented in Table~\ref{tab:metrics-emod}.

\paragraph{Experiment \expt{3}}
As shown in Table~\ref{tab:optimal-k}, the optimal \(k\) values for initializing EnsembleGaze varied across the object-level features. However, to ensure consistency across user-clustering configurations, a uniform value of \(k=3\) was applied throughout Experiment \expt{3}. Figure~\ref{fig:e3_emod} demonstrates that cluster membership varied significantly across all feature vector variations, with a relatively higher correspondence observed between configurations using all features and those using intensity-based features. While the coassociation matrix for intensity-based features displays the clearest structural patterns, it also appears to contain more than three distinct clustering partitions. Furthermore, compared to the consensus user-clustering of Experiment \expt{2}, the metrics in Table~\ref{tab:metrics-emod} reveal only average correspondence, characterized by low ARI and NMI values.

\paragraph{Experiment \expt{4}}
The stimulus-guided high-dimensional clustering configurations yielded a higher CH index for \(k\)-values of \((2,2)\), while simultaneously exhibiting suboptimal silhouette scores and DB indices. Figure~\ref{fig:e4_emod}A illustrates the correspondence between the stimulus-guided biclustering and consensus subspace clustering (CSC) schemes. The two configurations produce noticeably different partition sizes; specifically, CSC yields one large and one small partition, whereas biclustering generates two nearly equally-sized partitions. Bicluster C0 is composed primarily of members from CSC C1 alongside a subset from CSC C0, whereas bicluster C1 is almost entirely comprised of members from CSC C0. Compared to the Experiment \expt{2} consensus stimuli clustering, consensus partition C0 merges into CSC C1, while consensus C1 merges into CSC C0. The membership of consensus clusters C2 and C3 varies between the two CSC clusters, as shown in the first plot of Figure~\ref{fig:e4_emod}F. Additionally, the reordered activation heatmap (Figure~\ref{fig:e4_emod}G) exhibits a clear checkerboard pattern, indicating consistent within-group behavior among the stimuli partitions.

Further examination of the user partitions generated by these stimulus-guided configurations (Figure~\ref{fig:e4_emod}C) reveals that the biclustering user partitions all favor a single cluster, whereas the CSC scheme produces a fourteen-to-two divide. Comparing these results to the alluvial ribbons of the Experiment \expt{2} consensus user clustering, two consensus partitions (C1 and C2) merge entirely into CSC C0, while only a single member shifts from consensus C0 to CSC C0. Because the stimulus-guided sequential CSC demonstrated a higher CH index than the biclustering scheme, the composition of its two partitions is visualized in terms of color and intensity in Figure~\ref{fig:e4_emod}F. Furthermore, the feature composition charts for these configurations are presented in Figures~\ref{fig:e4_emod}H and~\ref{fig:e4_emod}I. Bicluster C0 is driven primarily by fixation duration features (\(\mu(FD)\), \(\sigma^2(FD)\), and \(\sum(FD)\)), whereas C1 is predominantly driven by \(\sum(FC)\). In contrast, CSC exhibits a comparatively high variance in fixation duration (\(\sigma^2(FD)\)) for the smaller partition C1, indicating that this metric serves as a distinct feature defining the cluster.

The user-guided high-dimensional clustering schemes do not yield optimal silhouette scores, DB indices, or CH indices for \(k\)-values of \((2,2)\). However, EnsembleGaze was initialized with these values to maintain uniformity across configurations. Figure~\ref{fig:e4_emod}B displays the correspondence between the Experiment \expt{2} consensus user clustering and the user-guided high-dimensional clustering schemes. The partition sizes produced by biclustering and CSC differ noticeably: biclustering produces evenly-sized partitions, whereas CSC generates one large and one small partition. The membership of seven users varies between the CSC and bicluster partitions. However, compared to the Experiment \expt{2} partitions, consensus C0 completely merges into CSC C0, while the single-member consensus cluster C2 merges into CSC C1. Conversely, the membership of consensus C1 is split between both CSC partitions.

As a result of these shifts, the overall correspondence metrics in Table~\ref{tab:metrics-emod} show very low Cohen's \(\kappa \), MCC, ARI, and NMI values for the user-guided high-dimensional clustering. This lack of clear structure is further supported by the reordered activation heatmap (Figure~\ref{fig:e4_emod}E), which fails to display a clear checkerboard pattern compared to previous configurations. Finally, examining the stimuli partitions of the user-guided configurations (Figure~\ref{fig:e4_emod}D) reveals a similar partition size mismatch between the CSC and biclustering outcomes. Compared to the consensus stimuli clustering of Experiment \expt{2}, two clusters (consensus C1 and C2) merge into CSC C0 with minimal crossover, whereas partition CSC C1 comprises a minority of stimuli that split away from the primary flows of their original partitions.

\section{Discussion}\label{sec:discussion}
\paragraph{Interpreting Clustering Quality}
While the silhouette score, DB index and CH index were used to evaluate the optimal $k$ value,we also use the same metrics to interpret the results and indicate the clustering quality. For example, all the distance-based user-clustering schemes of MIT1003 yielded very low silhouette scores and CH indices, and hence,  the partitions formed by these schemes are not very separable. This lack of separation is evidenced by indistinct block-diagonal patterns in the coassociation matrices, with the exception of the Hausdorff distance metric. Since the Hausdorff distance metric considers the points of a scanpath as a set, \ie~as a bag-of-points rather than as ordered sequences, it is able to converge on a lower value of $k$, while also exhibiting relatively better silhouette score, DB index and CH index. This conclusion suggests that the underlying user behavior of stable user groups may not require a strict one-to-one scanpath correspondence.

The consensus user-clustering and consensus stimuli-clustering of \expt{2} yielded relatively better silhouette score, DB index and CH index, indicating that the $k$ value chosen revealed separable clustering. Among the baseline clustering methods, MIT1003 aligned well with the spectral clustering outcomes, while EMOd aligned with that of KMeans++. This divergence demonstrates that the nature of the MIT1003 dataset's fixation patterns is non-linearly separable, whereas EMOd exhibited fixation patterns that were linearly separable. 

In the stimulus-guided, high-dimensional clustering of the MIT1003 and EMOd datasets, we observe a paradoxical effect: a high CH index occurs alongside a suboptimal DB index and silhouette score. Hence, we interpret the results of \expt{4} with caution. For MIT1003, the cluster correspondence between biclustering and CSC is high, revealing that both methods uncover stimuli groups subjected to similar fixation patterns. Given the lower silhouette score, the cluster centroids of the two partitions are likely spaced closely together. This proximity explains the fringe of stimuli flowing between partitions across both schemes, as illustrated in the alluvial diagram  (refer to alluvial diagram in Figure~\ref{fig:e4_mit}A). Conversely, on the EMOd dataset, the partitions of the two schemes do not align, resulting in low F1-scores.  The corresponding alluvial chart (Figure~\ref{fig:e4_emod}A) shows significant differences between the two schemes' partition sizes and the flow of members. Given the lower silhouette score here, we infer that these clusters are so tightly spaced that they lack even the relative separability observed in the MIT1003 partitions. 

Regarding the user-guided, high-dimensional clustering of the two datasets, we observe that the silhouette scores and DB indices are slightly better than their stimulus-guided counterpart, which presents their case more strongly. In the case of MIT1003, the user-guided configurations effectively reveal very similar partitions, with only one user changing membership between the two methods. The high Cohen's $\kappa$, MCC and F1 score also indicate strong alignment between the results, demonstrating clear separation. On the other hand, EMOd's user-guided high-dimensional clustering methods reveal very different partitions of varying sizes, and low correspondence metrics. This dissonance suggests one of two things: either the cluster centroids are too close to allow for distinct separation, or user partitions simply do not exist based on fixation features alone. This pattern is also observed in the user partitions of the stimulus-guided high-dimensional clustering of EMOd (Figure~\ref{fig:e4_emod}C), where all the users map to the same bicluster partition. Additionally, because the reordered activation heatmap of the user-guided biclustering lacks a strong checkerboard pattern, we infer that users may not possess separable characteristics based on fixations alone. Exploration of other features --- such as saccades, dwell time, AOI-based features --- is out of scope of this work and could be considered in the future to understand such complex dataset semantics.

\paragraph{Effect of Image-level and Object-level Properties}
While the MIT1003 dataset features predominantly red-hued images, EMOd contains a higher proportion of blue-hued images. Furthermore, the object detection model identified a larger share of ``People'' in EMOd compared to ``Objects'', ``Vehicles'', or ``Wildlife''. A significant portion of these detected objects were blue-hued with medium color intensity. In contrast, MIT1003 contains a high proportion of ``Objects'', ``Vehicles'', and ``People'', featuring a diverse mix of both red- and blue-hued objects.

These image-level color hues and intensities were used to construct composition charts for the consensus clusters across the different clustering configurations. Across all alluvial plots and composition charts, we find no strong evidence to support the hypothesis that fixations are guided by an image's color hue or intensity. This lack of a relationship is further affirmed by the low Cohen's \(\kappa \) and MCC values obtained for both datasets.

Conversely, our findings suggest that utilizing object-property-based fixations as features yields more robust clustering partitions. As shown in Figure~\ref{fig:e3_mit}, the coassociation matrices for MIT1003 exhibit distinct block-diagonal patterns, indicating the presence of cohesive user groups. Additionally, the cluster correspondence metrics between the consensus user-clustering of \expt{2} and \expt{3} display above-average values, providing strong evidence that object-based feature engineering reveals significant clustering. Similarly, Figure~\ref{fig:e3_emod} shows that the object-intensity-based coassociation matrix for EMOd displays a clear block-diagonal pattern. However, because the user-clustering configurations for EMOd fail to achieve high ARI and NMI values, the object-intensity-based user-clustering is only able to reach average values for Cohen's \(\kappa \) and MCC.

\paragraph{Features that Drive High-Dimensional Clusters}
Interestingly, the features that drive stimulus-guided, high-dimensional clustering partitions are clearly distinguishable in both datasets. Examining Figures~\ref{fig:e4_mit}H and ~\ref{fig:e4_mit}I reveals that the larger partition (partition 0) is driven by high values of total fixation count \(\sum{(FC)}\). Conversely, the smaller partition contains stimuli characterized by higher values of \(\sigma^2({FD}), \mu({FD}), \sum({FD})\) and \(TTF\). This indicates that a larger number of images received a higher count of fixations, suggesting that users did not allocate highly concentrated, focal attention to any single image. In other words, this larger stimulus partition received predominantly ambient attention. This aligns with existing literature on free-viewing tasks involving 2D natural scenery stimuli, which typically elicit initial ambient attention before resolving into progressively longer fixation durations indicative of focal attention \cite{valtchanov2015cognitive,henderson2018meaning}.

Regarding EMOd, we approach our inferences with caution due to the paradoxical behavior of the cluster quality indices. Because the features driving the biclustering and CSC partitions differ, we interpret them in conjunction with other clustering metrics. The biclustering partitions are nearly equal in size, and, partition 1 is driven by the total fixation count \(\sum{(FC)}\), suggesting a group of ambient-attention stimuli. In contrast, the CSC partitions are driven by high fixation duration variance \(\sigma^2{(FD)}\). Specifically, the smaller CSC partition (partition 1) shows very high variance in fixation duration, whereas the larger partition exhibits equivalent levels across all features. This suggests that images in partition 1 elicited a highly erratic fixation pattern composed of both short and long fixations. Such a pattern indicates moments of ambient attention interspersed with cognitive overload or confusion, where users felt compelled to repeatedly re-evaluate the stimuli. Given that EMOd contains images explicitly selected to evoke emotional responses, this finding is consistent with literature on distressing stimuli~\cite{calvo2004gaze,castellanos2026effects}. Contrasting the biclustering and CSC results reveals contradictory trends: biclustering fails to discover the erratic attention pattern, while CSC fails to isolate the ambient-attention driven stimuli. However, because stimulus-guided biclustering achieved lower silhouette scores and CH indices than CSC, we conclude that the stimulus-guided CSC partitions offer a more robust representation of the features driving EMOd clustering.

\section{Conclusion}\label{sec:conclusion}
We presented an end-to-end system, EnsembleGaze, for the systematic investigation of unsupervised learning strategies for clustering of fixation-based gaze data, involving consensus clustering and high-dimensional clustering. The high-dimensional clustering itself included spectral biclustering using two different matrix unfolding techniques, and consensus subspace clustering using two different sequence of dimensions for primary and secondary clusters. All the clustering methods depend on the distributional statistics of fixation feature vectors, and to a lesser extent, scanpath-distance-based features.

Evaluating the clustering quality across two case studies demonstrates that while any algorithm can generate a set of \(k\) partitions, meaningful interpretation requires analyzing these results alongside cluster quality indices and correspondence metrics. Utilizing consensus clustering established a robust baseline for evaluating all subsequent configurations. Through this framework, we found that image-level color properties do not strongly dictate user fixation patterns, while, object-property-based feature engineering offers a robust method to discover distinct user-clustering partitions. This is affirmed by the distinct block-diagonal patterns observed in the coassociation matrices, which prove that underlying user behavior can be effectively mapped into cohesive, stable groups. Furthermore, the choice of distance metric and clustering configuration plays a pivotal role in revealing these group semantics, as evidenced by the unique, set-based convergence of the Hausdorff metric. The analysis also uncovers a clear distinction between ambient and focal attention mechanisms in the MIT1003 dataset, while identifying erratic fixation duration variances triggered by the emotion-eliciting stimuli in EMOd. In this manner, consensus subspace clustering successfully complements the spectral biclustering method. Remarkably, we note that while fixation behavior effectively reveals visual attention patterns, certain configurations yield non-separable cluster partitions. This ambiguity requires further investigation to determine whether non-separability is an inherent property of the dataset itself, or if incorporating alternative features would generate more separable partitions.
The present study confines its analysis to fixation-based features, leaving the complementary role of saccadic behavior unexplored. Including saccade-based features in the analysis could reveal whether the observed clusters correspond to distinct attentional patterns, for instance distinguishing ambient scanning patterns, which is characterized by broad, exploratory eye movements, from focal patterns, which are marked by sustained, localized attention. A further limitation concerns the reliance of EnsembleGaze on a fixed set of hand-engineered features for each clustering scheme. An important direction for future work is the development of data-driven feature learning approaches that could automatically identify the features that are most discriminative of each clustering partition, potentially uncovering structure not captured by the current engineered features. Recent advances in transformers and learning methods could enable feature learning. Another future work could investigate sequence-aware representations of gaze behavior, where the ordered sequence of fixations and saccades in a scanpath could be considered as a feature. Such temporal features would better capture viewing strategies like task-oriented vs. task-free viewing and may enable more accurate, automated user grouping. Finally, as an alternative to unsupervised learning, community detection methods can be explored, as it could uncover complex structural relationships, with no \textit{a priori} initialization of the number of clusters $k$. 

\clearpage
\bibliographystyle{unsrt}
\bibliography{papers_eyetracking}
\end{document}